%% file: main.tex
\documentclass[10pt,twocolumn,letterpaper]{article}

\usepackage[pagenumbers]{cvpr} 

\input{preamble}

\definecolor{cvprblue}{rgb}{0.21,0.49,0.74}
\usepackage[pagebackref,breaklinks,colorlinks,citecolor=cvprblue]{hyperref}

\usepackage{tabularx}
\usepackage{color}
\usepackage{colortbl}
\usepackage{xcolor}
\usepackage{multirow}
\usepackage{multicol}
\usepackage{wrapfig}
\usepackage{marvosym}

\title{CoHD: A Counting-Aware Hierarchical Decoding Framework 
\\ for Generalized Referring Expression Segmentation}

\author{Zhuoyan Luo$^{1}$\footnotemark[2], \quad Yinghao Wu$^{1}$\footnotemark[2],\\
  Tianheng Cheng$^{2}$, \hfill 
  Yong Liu$^{1}$, \hfill 
  Yicheng Xiao$^{1}$, \hfill
  Hongfa Wang$^{3}$, \hfill
  Xiao-Ping Zhang$^{1\textrm{\Letter}}$, \hfill
  Yujiu Yang$^{1\textrm{\Letter}}$ \\
  $^{1}$Tsinghua Shenzhen International Graduate School, Tsinghua University\\
  $^{2}$School of EIC, Huazhong University of Science and Technology \quad $^{3}$Tencent TEG\\
  {\tt\small \{luozy23, yh-wu23\}@mails.tsinghua.edu.cn} \\
}

\begin{document}
\maketitle
\renewcommand{\thefootnote}{\fnsymbol{footnote}}
\footnotetext[2]{Equal contribution.}
\footnotetext[0]{${\textrm{\Letter}}$ Corresponding author.}
\input{sec/0_abstract}
\input{sec/1_intro}
\input{sec/2_related}
\input{sec/3_method}
\input{sec/4_experiment}

\input{sec/5_conclusion}

{
    \small
    \bibliographystyle{ieeenat_fullname}
    \bibliography{main}
}

\input{sec/appendix}

\end{document}



\input{sec/X_suppl}

{
    \small
    \bibliographystyle{ieeenat_fullname}
    \bibliography{main}
}

%% file: preamble.tex
\usepackage[dvipsnames]{xcolor}


%% file: sec/0_abstract.tex
\begin{abstract}
The newly proposed Generalized Referring Expression Segmentation (GRES) amplifies the formulation of classic RES by involving complex multiple/non-target scenarios.
Recent approaches address GRES by directly extending the well-adopted RES frameworks with object-existence identification.
However, these approaches tend to encode multi-granularity object information into a single representation, which makes it difficult to precisely represent comprehensive objects of different granularity.
Moreover, the simple binary object-existence identification across all referent scenarios fails to specify their inherent differences, incurring ambiguity in object understanding.
To tackle the above issues, we propose a \textbf{Co}unting-Aware \textbf{H}ierarchical \textbf{D}ecoding framework (CoHD) for GRES.
By decoupling the intricate referring semantics into different granularity with a visual-linguistic hierarchy, and dynamic aggregating it with intra- and inter-selection, CoHD boosts multi-granularity comprehension with the reciprocal benefit of the hierarchical nature.  
Furthermore,
we incorporate the counting ability by embodying multiple/single/non-target scenarios into count- and category-level supervision, facilitating comprehensive object perception.
Experimental results on gRefCOCO, Ref-ZOM, R-RefCOCO, and RefCOCO benchmarks demonstrate the effectiveness and rationality of CoHD which outperforms state-of-the-art GRES methods by a remarkable margin. Code is available at \href{https://github.com/RobertLuo1/CoHD}{here}.
\end{abstract}

%% file: sec/1_intro.tex
\section{Introduction}
\label{sec:intro}

\begin{figure}[ht]
    \centering
    \includegraphics[width=\linewidth]{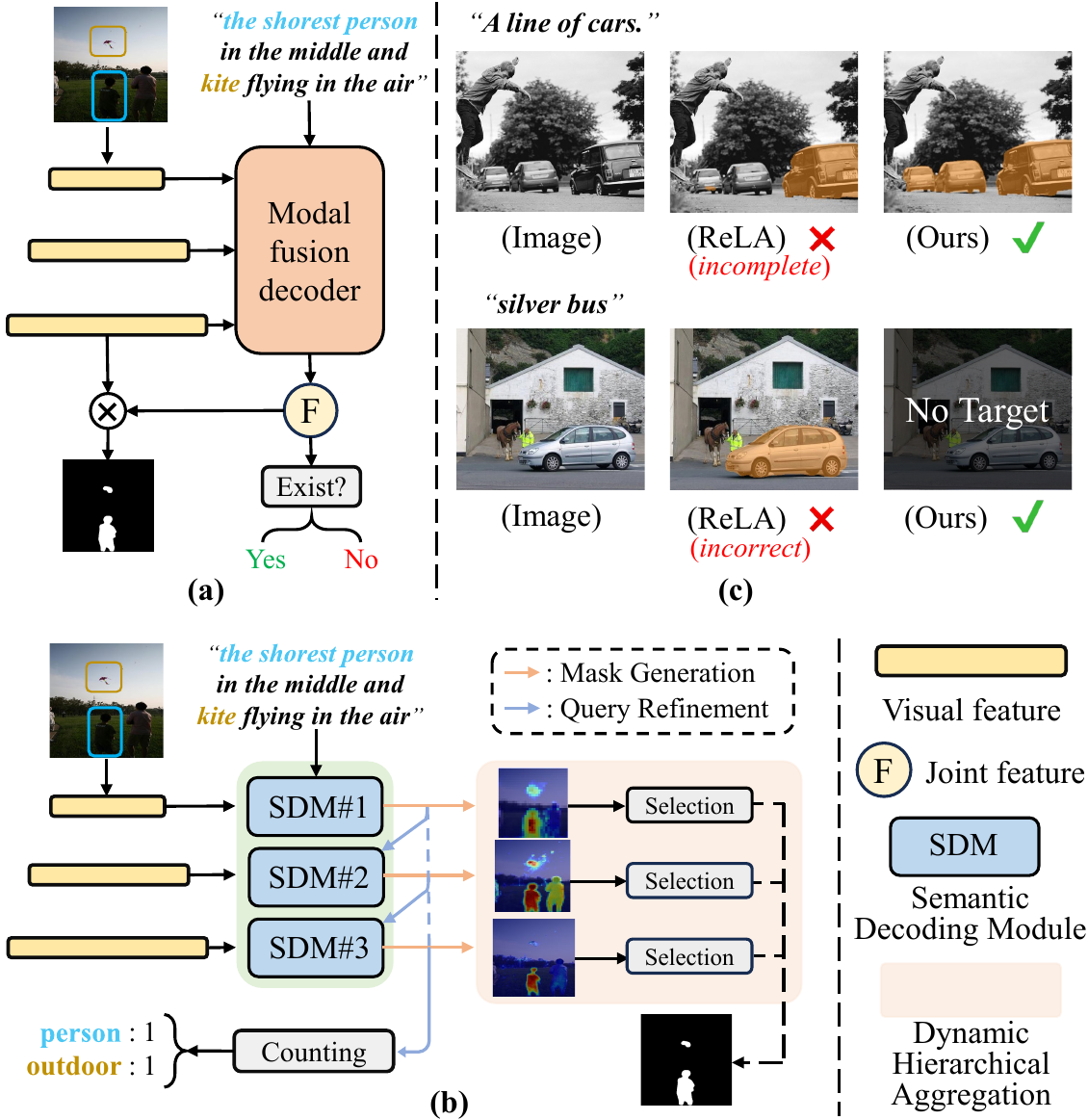}
    \vspace{-20pt}
    \caption{\textbf{Comparison of decoding paradigms.} Previous GRES methods attempt to integrate multi-granularity context into one joint feature for mask and object-existence prediction (a), which exhibits vulnerability to the complexity of multiple/non-target scenarios (c). 
    Instead of relying on the single representation, we decouple the referring correspondence into the semantic hierarchy of different granularity and perform dynamic selective aggregation for robust segmentation.  
    Further, in contrast to simple binary classification, we empower CoHD with counting ability to promote object perception (b).}
    \label{fig:teaser}
    \vspace{-15pt}
\end{figure}

Referring Expression Segmentation (RES) is one of the fundamental tasks in visual-linguistic understanding, which aims at localizing the referred object in an image with a natural language description. Despite achieving great progress in recent years~\cite{husegmentation2016, lavt, vlt, mmm, cris}, the simplification of RES formulation, where one object must be included in a sentence, still constrains its potential in various scenarios. Consequently, such limitation encourages the emergence of Generalized Referring Expression Segmentation (GRES). Different from the one-to-one setting in classic RES, GRES supports extra multiple-target and empty-target cases, rendering it more suitable in real-world applications.

However, it poses new challenges with the additional generalized settings. The multiple-target referent necessitates more comprehensive semantic clues to handle complicated spatial-text relationships between instances for arbitrary location segmentation. On the other hand, the empty-target one highlights the significance of contextual information in mitigating undesirable responses from similar objects.  
An intuitive approach is to endow prevalent RES methods~\cite{lavt, vlt, efn} with multi/non-target referring capability.
For instance, ReLA~\cite{rela} introduces a large-scale generalized dataset gRefCOCO and utilizes spatial-aware queries to transform GRES into a related region retrieval task.
Meanwhile, DMMI~\cite{refzom} and RefSegformer~\cite{rris} enhance modality interaction in the early-fusion process and incorporate new benchmarks Ref-ZOM and R-RefCOCO respectively, to realize robust segmentation.

Despite advancements in GRES, the direct extension from RES to GRES exhibits less compatibility in two aspects:
1) \textbf{\textit{Incomplete Semantic Context}}: as shown in \cref{fig:teaser} (a) and (c), 
the previous decoding paradigm is solely dependent on a single representation for multi-granularity modeling, which leads to incomplete or incorrect object responses.  
We speculate that primarily stems from the increased complexity of referring semantics in GRES, \textit{e.g.}, intricate spatial relationships among instances, and deceptive expressions. It easily incurs the shifted emphasis on both visual and language-wise, thereby impeding the integration of all-grained contexts into one joint representation.   
2) \textbf{\textit{Imprecise Object Perception}}: previous GRES methods adopt simple binary classification for object-existence identification, which ignores the inherent distinction between multiple and single referent scenarios, potentially incurring ambiguity in object understanding.

To well address the challenges of GRES and break through the limitations of previous methods, we delve into a further investigation. Firstly, the extra referring scenarios, \textit{i.e.}, multiple/empty-target referents, bring more difficulty in accurately activating complete regions and discerning the non-related ones. For instance, as shown in \cref{fig:teaser} (b), different referring object activations vary in shapes, scales, and locations, and misinterpretation potentially occurs at the latter level. It indicates that simply relying on a single joint feature for mask decoding is insufficient. Instead, the complex referring semantics should be decoupled by different granularity and aggregated hierarchically for robust segmentation.  
Secondly, the inherent distinction of multiple/single referent cases should not be ignored since it makes a difference to the precision object perception when encountering implicit expressions such as ``\textit{A line of cars}''. 
Inspired by that, we present a \textbf{Co}unting-aware \textbf{H}ierarchical \textbf{D}ecoding framework (CoHD). Specifically, a multi-grained visual-linguistic hierarchy is formulated by the Hierarchical Semantic Decoder (HSD). Instead of simply leveraging a single joint feature, HSD performs dynamic hierarchical aggregation with intra- and inter-selection. 
Thus, the intricate referring semantic clues can be completely leveraged. 
Furthermore, we introduce the Adaptive Object Counting module (AOC) to specify the inherent distinction among multiple/single/non-target scenarios by embodying each into count- and category-level supervision, which tackles ambiguity and facilitates precise object perception.  

We conduct extensive experiments on three popular GRES benchmarks, \textit{i.e.}, gRefCOCO~\cite{rela}, Ref-ZOM~\cite{refzom} and R-RefCOCO~\cite{rris} along with RES datasets RefCOCO/+/g~\cite{refcoco, refcocog}. 
Experimental results demonstrate that CoHD remarkably outperforms previous works on all benchmarks, showcasing the effectiveness of the proposed method for complicated scenarios in GRES.

Overall, our contributions are summarized as follows:
\begin{itemize}
    \item We present the Counting-Aware Hierarchical Decoding framework for GRES, aiming to decouple the complex referring semantics into different levels for boosting multi-granularity comprehension. In CoHD, the Hierarchical Semantic Decoder (HSD) formulates a visual-linguistic hierarchy of different granularity and performs mask decoding by dynamic hierarchical aggregation.  
    \item We introduce the Adaptive Object Counting module (AOC) to facilitate precise object perception by assigning explicit counting capability, which seamlessly adapts to the specificity of multiple/single/non-target scenarios.
    \item Our approach outperforms all existing state-of-the-art methods by a remarkable margin without bells and whistles, \textit{e.g.}, $+4.8\%$ gIoU in gRefCOCO, $+7.5\%$ rIoU in R-RefCOCO and $+6.3\%$ Acc. in Ref-ZOM. 
\end{itemize}

%% file: sec/2_related.tex
\section{Related Work}
\label{sec:related_work}

\begin{figure*}[t]
    \centering
    \vspace{-5pt}
    \includegraphics[width=\linewidth]{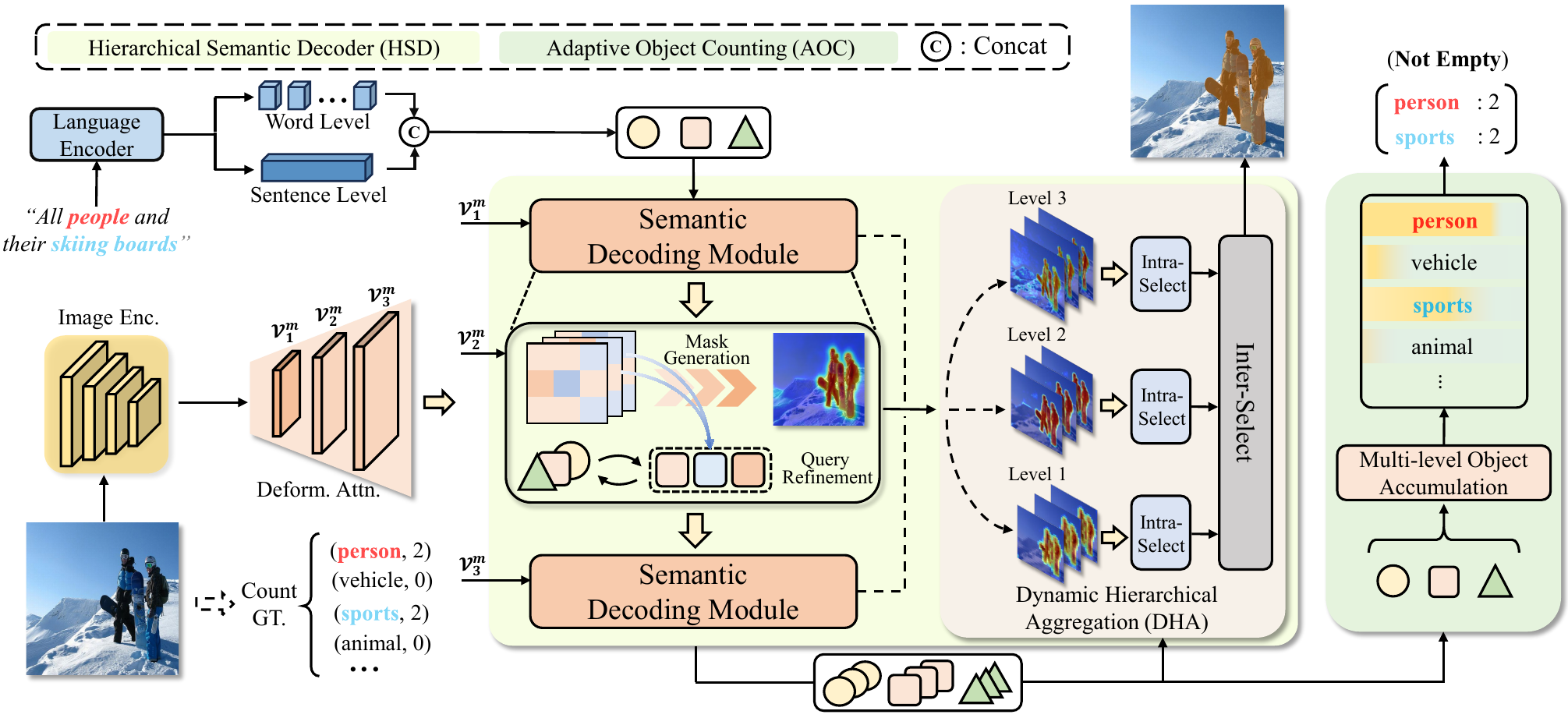}
    \vspace{-20pt}
    \caption{\textbf{Overview of CoHD}. The model takes an image with the referring expression as input. After the encoding process, 
    the Hierarchical Semantic Decoder (HSD) decouples the referring semantics into different granularity for establishing visual-linguistic hierarchy and then performs dynamic aggregation with intra- and inter-selection for final segmentation. 
    Moreover, with well-aligned semantic contexts, we introduce the Adaptive Object Counting module (AOC) to fully promote comprehensive object perception with the category-specific counting ability under multiple/single/non-target scenarios.}
    \label{fig:framework}
    \vspace{-10pt}
\end{figure*}

\paragraph{Referring Expression Segmentation.} Referring Expression Segmentation (RES) aims to localize a single object referred by a natural language description, which is first introduced by Hu \textit{et al.}~\cite{husegmentation2016}. Conventional approaches~\cite{husegmentation2016, liu2017, chensee2019, cmpc, efn} typically utilize convolution-based operations as the cross-modal fusion to generate segmentation masks. To overcome the weakness of insufficiency in modeling visual-text relation that prior works suffer, recent studies~\cite{lavt,vlt,mmm,cris,Restr} design advanced attention-based mechanism~\cite{transformer} to facilitate multi-modal interaction. 
Since previous works are restricted by pre-defined rules that one sentence must match an object, ReLA~\cite{rela} proposes the Generalized Referring Expression Segmentation (GRES) task, which additionally involves empty-target and multiple-target scenarios. DMMI~\cite{refzom} introduces a new benchmark and baseline to realize beyond one target segmentation. Similarly, RefSegformer~\cite{rris} endows the transformer-based model with empty-target sentence discrimination.
Further, GSVA~\cite{gsva} reuses the [SEG] token and additionally incorporates the [REJ] one into LISA~\cite{lisa}, which equips the MLLM with GRES ability.
However, these methods seek to encode multi-granularity object information into a single joint representation for mask prediction, which is impractical in GRES. In contrast, we aggregate visual-linguistic semantic hierarchy of different granularity in a hierarchical manner for more generalized segmentation.    

\paragraph{Object Counting.} Object counting is to predict the number of conditional objects within an image, which can be divided into detection-based and regression-based approaches. The detection-based methods~\cite{psc,goldman2019,heish2017} obtain the number of instances by proposals localization and ranking while the regression-based ones~\cite{cholakkal2019, can, sun2023} generates predicted counts via density map integration. Different from the classic object counting mechanism, our proposed Adaptive Object Counting module performs at the query level. By considering various scenarios, \textit{i.e.}, multiple/single/non-target referents, our model achieves comprehensive object perception under each situation. 

%% file: sec/3_method.tex
\section{Method}
\subsection{Overview}
\begin{figure*}[t]
    \centering
    \includegraphics[width=\linewidth]{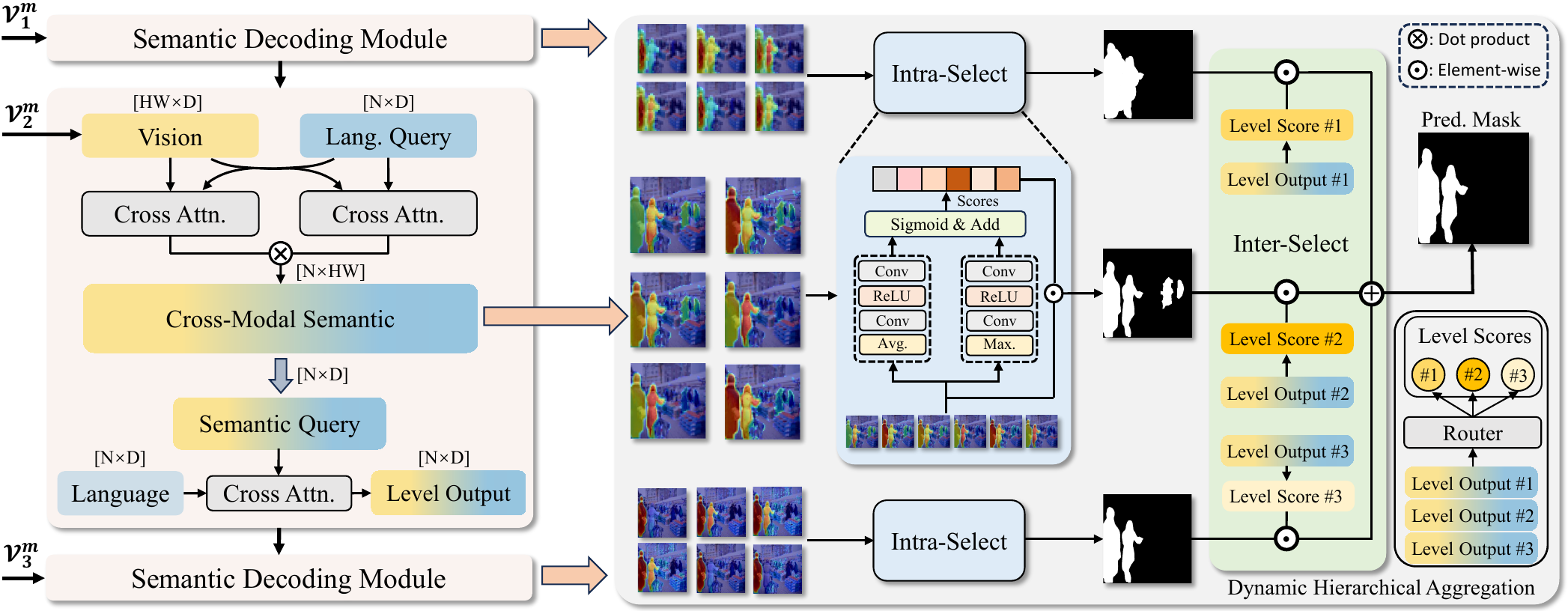}
    \vspace{-15pt}
    \caption{\textbf{Illustration of Hierarchical Semantic Decoder.} The Semantic Decoding Module takes visual and linguistic features as input and enables bi-directional modality calibration to generate the corresponding fine-grained semantic map at each level, which formulates a visual-linguist hierarchy. Subsequently, we incorporate the referring semantics into the query level with the original language information, which facilitates a more holistic object understanding. With the well-aligned semantic hierarchy, we perform dynamic aggregation with the intra- and inter-selection mechanism for mask decoding to excavate the potential of reciprocal benefits brought by the hierarchical nature.}
    \label{fig:decoder}
    \vspace{-10pt}
\end{figure*}

\cref{fig:framework} depicts the overall pipeline of CoHD. Given an image $\mathcal{I}$, the visual backbone, \textit{e.g.}, Swin-Base~\cite{swin} is to extract a set of vision features $\mathcal{V}_i\in \mathbb{R}^{D^{V} \times H_i \times W_i}$, $i\in \{1, 2, 3, 4\}$, where $H_i$ and $W_i$ denote the width and height of each scale of feature maps respectively. $D^{V}$ signifies the channel dimension. $\mathcal{V}_i$ is further encoded as mask features $\mathcal{V}^m_i$ ($i\in \{1, 2, 3\}$) by deformable attention~\cite{deformabledetr}. Simultaneously, the
 textual expression $\mathcal{E}$ with $T$ words is sent into a transformer-based~\cite{bert} language encoder which generates the word-level embedding $\mathcal{L}^w \in \mathbb{R}^{T \times D^{L}}$ and sentence-level embedding $\mathcal{L}^s \in \mathbb{R}^{D^{L}}$. Considering that $\mathcal{L}^w$ contains fine-grained descriptions while $\mathcal{L}^s$ expresses the general information, language query $\mathcal{L}^O \in \mathbb{R}^{N \times D^{L}}$ is obtained via the concatenation of them. $\mathcal{L}^O$ and $\mathcal{V}^m_i$ are simultaneously sent into the Hierarchical Semantic Decoder (HSD), which generates corresponding object segmentation $\mathcal{M} \in \mathbb{R}^{H \times W}$ 
 by dynamic hierarchical mask decoding on the semantic hierarchy of different granularity. Subsequently, Adaptive Object Counting (AOC) is designed to promote the precise object perception of different scenarios by counting ability. 

\subsection{Hierarchical Semantic Decoder}
Previous GRES methods~\cite{rela, refzom, rris} adopt the traditional RES decoding paradigm which only incorporates the last modality-fused features. It is acceptable in RES since it is to \textbf{\textit{match}} the single referred object through multi-modal interaction. However, it is less compatible with GRES. We observe that misinterpretation of region activation potentially occurs at the specific granularity in multiple-target cases, which indicates that simply performing sequential matching within each level as RES will lead to incomplete object segmentation. For the non-target referent, the final joint representation of multi-grained information is highly redundant, possibly resulting in incorrect object responses. Therefore, the hierarchical \textbf{\textit{aggregation}} across granularity is significant for GRES, since it benefits from the reciprocal properties of semantic activation at each granularity, which helps to address the previous issues. 
Inspired by that, we propose a Hierarchical Semantic Decoder to formulate a visual-linguistic hierarchy through concatenated semantic decoding modules and perform hierarchical dynamic aggregation for robust segmentation.

\paragraph{Semantic Decoding Module.}
The core idea of SDM is in two folds: 1) \textit{Mask Generation}: generate the fine-grained semantic map at each scale of visual features which is further utilized in the dynamic aggregation process. 2) \textit{Query Refinement}: the well-supervised semantic map helps guide the formulation of holistic comprehension of the referring semantics at the query level. Thus, the intricate semantics are well modeled by the mutual-enhanced relationship within modules.

Specifically, taking $\mathcal{V}^m_1$ as an example, we firstly leverage fully connected layers to map visual features and language query into a shared joint space $D$, respectively. The mapped representations are further denoted as $\mathcal{V}$ and $\mathcal{L}$ for clarity.
Then, we derive the coarse semantic map $\mathcal{A}$ by attention mechanism and it is applied in visual-linguistic interaction to align information flow between modalities for generating the fine-grained semantic map $\mathcal{S}$. 
The modality complementary highlights the corresponding object activation of language-guided visual features, while simultaneously relieving the potential noise in visual-oriented textual representations.
The above process can be formulated as:  
\begin{gather}
    \mathcal{A} =\frac{1}{\sqrt{D}}\left(\mathcal{L}\mathcal{W}^{k}_{L}\right)\left(\mathcal{V}\mathcal{W}^{k}_{V}\right)^T, \\
    \mathcal{F}^{L \sim V} = \mathrm{softmax}(\mathcal{A}) \times (\mathcal{V}\mathcal{W}^{v}_{V}), \\
    \mathcal{F}^{V \sim L} = \mathrm{softmax}(\mathcal{A}^T) \times \left(\mathcal{L}\mathcal{W}^{v}_{L}\right), \\
    \mathcal{S} = \mathcal{F}^{L \sim V} \times (\mathcal{F}^{V\sim L})^T,
\end{gather}
where $\mathcal{W}^{k}_{V}$, $\mathcal{W}^{v}_{V}$, $\mathcal{W}^{k}_{L}$, $\mathcal{W}^{v}_{L}$ are learnable weights that map visual and linguistic representations into attention space. 
In this manner, the key tokens related to the referent are anticipated to emerge significant responses in the semantic map $\mathcal{S}$ while suppressing the activation of non-related regions. Then, it is further utilized as the input $\mathcal{M}_1 \in \mathbb{R}^{N \times H_1W_1}$ to formulate the multi-granularity visual-linguistic correspondence hierarchy for dynamic aggregation. 

Considering that the textual description in GRES weighs more crucial than the classic RES task since it encompasses additionally implicit multiple-target information or deceptive description, we further introduce the well-aligned semantics into the query level for better object perception (see in \cref{fig:decoder}). It well incorporates the important referring semantics and preserves valuable language information simultaneously.
Formally, we leverage fully connected layers to project the fine-grained semantic map $\mathcal{S}$ into query space:
\begin{equation}
    \mathcal{Q}^{S} = \mathrm{MLP}(\mathcal{S}).
\end{equation}
The high-level semantics in $\mathcal{Q}^{S}$, instead of the single visual modality, is injected into the language query $\mathcal{Q}^{L}$, \textit{i.e.}, the initial input query $\mathcal{L}^{O}$ or the one from previous SD module:
\begin{equation}
    \mathcal{Q}^{L'} = \mathrm{MHA}\left(\mathcal{Q}^{L}, \mathcal{Q}^{S}, \mathcal{Q}^{S}\right),
\end{equation}
where $\mathrm{MHA}$ is the vanilla multi-head attention mechanism and $\mathcal{Q}^{L'} \in \mathbb{R}^{N \times D}$. With greatly enriched by the referring semantics from $\mathcal{Q}^{S}$, we further perform language reactivation by reintroducing the original linguistic feature $ \mathcal{L}$, which not only prevents the shifted emphasis but also facilitates a more holistic understanding of that.

\paragraph{Dynamic Hierarchical Aggregation.}
We observe that the desired object responses typically vary in granularity when performing multiple object referring. Similar to the non-target one, the unrelated region activation will potentially occur due to the deceptive descriptions. Therefore, we emphasize the importance of reciprocal properties of multi-granularity information in GRES by proposing Dynamic Hierarchical Aggregation to unify the visual-linguistic hierarchy with intra- and inter-selective mechanisms.
Shown in \cref{fig:decoder}, the output $ \{\left(\mathcal{M}_1, \mathcal{Q}^{L'}_{1}\right), \left(\mathcal{M}_2, \mathcal{Q}_{2}^{L'}\right),
\left(\mathcal{M}_3, \mathcal{Q}_{3}^{L'}\right)\}$ from cascaded SDMs are gathered into DHA for segmentation. It is anticipated that $\mathcal{M}_1$ with the largest downsampling rate contains the highest level of object semantics, \textit{e.g.}, location, being referred or not, etc. In contrast, $\mathcal{M}_3$ mainly perceives the shape, and texture character of the objects. However, simply aggregation by summation easily introduces bias accumulation, leading to unsatisfactory outcomes. To fully excavate the benefit of hierarchical nature, we perform intra- and inter-selection on $\mathcal{M}$ and $\mathcal{Q}^{L'}$ respectively. 1) \textit{Inter-Selection}: inspired by the current success of MoE~\cite{moe}, we design a gating network to assign the saliency score for each level, which adaptively highlights the related object responses through granularity:
\begin{gather}
\alpha_i = \mathcal{G}(\mathcal{Q}_i^{L'}),  i \in \{1, 2, 3\}\\
\mathcal{G}(x):= \mathrm{Sigmoid}(x \mathcal{W}_g).
\end{gather}
where $\alpha$ $\in$ $\mathbb{R}^{1}$ and $\mathcal{W}_g$ is the learnable weights.
2) \textit{Intra-Selection}: affected by the deceptive or compound structure textual description, some tokens may exhibit sparse and cluttered information pertaining to targets in $\mathcal{M}$. We assume that the informative and highly related region activation plays a significant role in hierarchical aggregation. With the purpose of dynamically enhancing the referred object responses and simultaneously suppressing the non-related ones, we leverage channel attention module~\cite{channelattn} on $\mathcal{M}_{i}$ to evaluate the degree of referring correspondence in token level:
\begin{equation}
\mathcal{M}_{i}' = \mathrm{ChannelAttn}(\mathcal{M}_{i}) \cdot \mathcal{M}_{i}.
\end{equation}
After suppressing the unrelated activation based on the low referring correspondence, $ \{\mathcal{M}_1', \mathcal{M}_2', \mathcal{M}_3'\} $ are weighted aggregated with each corresponding saliency score $\alpha_{i}$ through sequential upsampling in a multi-level paradigm. The whole process can be expressed as:
\begin{equation}
  \left\{
  \begin{aligned}
   \mathcal{M}_{1}^* &= \alpha_1\mathcal{M}_{1}', \\
   \mathcal{M}_{i+1}^* &=  \mathrm{Up}(\mathcal{M}_i^*) + \alpha_{i+1}\mathcal{M}_{i+1}', \quad i \in \{1, 2\}
  \end{aligned}
  \right.
\end{equation}
Finally, we convert the aggregated semantic-enriched segmentation $\mathcal{M}^{*}_{3}$ into a two-class mask $\mathcal{M}^{o}$ by multiplying kernel $B$ which is conditional on the $\mathcal{Q}_{3}^{L'}$:
\begin{equation}
  \mathcal{M}^{o} = \mathcal{M}^{*}_{3} \times B.
\end{equation}

\input{table/grefcoco}
\input{table/rrefcoco}

\subsection{Adaptive Object Counting}

Existing GRES methods tend to directly adopt a binary classification branch to judge the existence of objects. However, we observe that such a simple design exhibits less compatibility for general referring segmentation settings. 
Particularly, it roughly categorizes multiple-target and single-target scenarios into one objective, ignoring their inherent specificity, which can easily incur ambiguity when encountering implicit or counting expressions such as \textit{``a line of cars"} or \textit{``two adult cow"}.  

Meanwhile, several studies~\cite{mmm, refzom} use the language reconstruction technique. Yet, it is not suitable in GRES since reconstructing unrelated content increases the burden of precise object awareness. 
Therefore, we propose Adaptive Object Counting (AOC), aiming to endow our model with category-specific counting ability in multiple/single/non-target situations. It significantly facilitates the precise understanding of both visual scenes and complicated sentences, \textit{i.e.}, counting, and compound structure expressions. 

Formally, the multi-modal semantic queries $\{ \mathcal{Q}^{L'}_1,\mathcal{Q}^{L'}_2,\mathcal{Q}^{L'}_3 \}$ corresponding to each granularity are transformed into the counting vector of $C$ categories:
\begin{equation}
    \mathcal{C}_i = \mathrm{MLP}(\mathcal{Q}^{L'}_i),
\end{equation}
where $\mathcal{C}_i \in \mathbb{R}^{N \times C}$. Then $\{ \mathcal{C}_1, \mathcal{C}_2, \mathcal{C}_3 \}$ are incorporated for the counting prediction $\mathcal{C}_{all} \in \mathbb{R}^{N \times C}$ by average pooling. Finally, we provide the explicit supervision on $\mathcal{C}_{all}$ by computing the $\mathit{L_1}$ distance between the multi-level accumulated results $\mathcal{C}^{pred} \in \mathbb{R}^{C}$ and the ground truth $\mathcal{C}^{gt}$, where $\mathcal{C}^{gt}$ (see in \cref{fig:framework}) is annotated as the exact counts of each category. The whole process can be formulated as:
\begin{equation}
    \begin{aligned}
    \mathit{Loss}_{count} &= \mathit{SmoothL1}(\mathcal{C}^{pred} - \mathcal{C}^{gt}), \\
    \mathit{SmoothL1}(x) &= 
    \begin{cases} 
     0.5 \cdot x^2 & \text{if } |x| < 1. \\
     |x| - 0.5 & \text{otherwise}.
    \end{cases}
    \end{aligned}
\end{equation}
On one hand, the AOC module is designed to seamlessly adapt to the specificity in generalized scenes, \textit{i.e.}, multiple/single/non-target tasks, which benefits the general object matching. 
On the other hand, guided by the the precise categories and counts of referent targets, our method facilitates comprehensive object perception in both image and language wise.

\subsection{Output and Loss}
We supervise our framework with three types of training objectives. 
Similar to~\cite{lavt, rela, refzom}, binary cross entropy loss $\mathit{Loss}_{mask}$ is adopted to measure the disparity between GT $\mathcal{M}^{gt}$ and the final mask prediction $\mathcal{M}^o$. As discussed above, $\mathit{Loss}_{count}$ is to promote the object perception. Due to the constrain of current GRES evaluation metrics, a lightweight count head is built upon AOC to judge the existence of objects. It is worth noting that it only trains on $\mathcal{C}^{pred}$ and will not have any impact on the main framework, which is highly different from existing GRES methods:
\begin{equation}
    \begin{gathered}
    \mathcal{P}^{pred} = \mathrm{MLP}\left(\mathrm{Detach}(\mathcal{C}^{pred})\right), \\
    \mathit{Loss}_{exist} = \mathit{Loss}_{ce}\left(\mathcal{P}^{pred}, \mathcal{P}^{gt} \right).
    \end{gathered}
\end{equation}
Therefore, total loss with balance factor $\lambda$ is:
\begin{equation}
    \mathit{Loss} = \lambda_{1}\mathit{Loss}_{mask} + 
 \lambda_{2}\mathit{Loss}_{count} + \lambda_{3}\mathit{Loss}_{exist}.
\end{equation}

%% file: table/grefcoco.tex
\begin{table*}
\begin{center}
\setlength{\tabcolsep}{4.5mm}{
\renewcommand\arraystretch{1.0}
\resizebox{\linewidth}{!}{
\begin{tabular}{l|c|ccc|ccc|ccc}
\specialrule{.1em}{.05em}{.05em}
\multirow{2}{*}{Method} & \multirow{2}{*}{Backbone} & \multicolumn{3}{c|}{Validation Set} & \multicolumn{3}{c|}{Test Set A}  & \multicolumn{3}{c}{Test Set B} \\
& & gIoU & cIoU & N-acc. & gIoU & cIoU & N-acc. & gIoU & cIoU & N-acc. \\
\midrule
\multicolumn{10}{c}{\textit{MLLM Methods}} \\
\midrule
LISA-V-7B~\cite{lisa} & SAM-ViT-H & 32.21 & 38.72 & 2.71 & 48.54 & 52.55 & 6.37 & 39.65 & 44.79 & 5.00 \\

GSVA-V-7B~\cite{gsva} & SAM-ViT-H & 63.32 & 61.70 & 56.45 & 70.11 & 69.23 & 63.50 & 61.34 & 60.26 & 58.42  \\
LISA-V-7B~\cite{lisa} (ft) & SAM-ViT-H & 61.63 & 61.76 & 54.67 & 66.27 & 68.50 & 50.01 & 58.84 & 60.63 & 51.91 \\

GSVA-V-7B~\cite{gsva} (ft) & SAM-ViT-H & 66.47 & 63.29 & 62.43 & 71.08 & 69.93 & 65.31 & 62.23 & 60.47 & 60.56 \\

\midrule
\multicolumn{10}{c}{\textit{Specialist Methods}} \\
\midrule
MattNet~\cite{mattnet} & ResNet-101 & 48.24 & 47.51 & 41.15 & 59.30 & 58.66 & 44.04 & 46.14 & 45.33 & 41.32 \\
LTS~\cite{lts} & DarkNet-53 & 52.70 & 52.30 & - & 62.64 & 61.87 & - & 50.42 & 49.96 & - \\
VLT~\cite{vlt} & DarkNet-53 & 52.00 & 52.51 & 47.17 & 63.20 & 62.19 & 48.74 & 50.88 & 50.52 & 47.82 \\
CRIS~\cite{cris} & CLIP-R101 & 56.27 & 55.34 & - & 63.42 & 63.82 & - & 51.79 & 51.04  & - \\

ReLA$^{\dagger}$~\cite{rela} & Swin-T & 56.87 & 57.73 & 44.07 & 65.29 & 66.02 & 48.68 & 57.37 & 56.86 & 50.31 \\
\rowcolor{gray!10} \textbf{CoHD} (Ours) & Swin-T &
\textbf{65.89} & \textbf{62.95} & \textbf{60.95} & \textbf{70.17} & \textbf{68.93} & \textbf{61.52} & \textbf{61.78} & \textbf{60.32} & \textbf{60.00} \\
\midrule
LAVT~\cite{lavt} & Swin-B & 58.40 & 57.64 & 49.32 & 65.90 & 65.32 & 49.25 & 55.83 & 55.04 & 48.46 \\
DMMI$^{*}$~\cite{refzom} & Swin-B & 62.68 & 62.78 & 53.20 & 68.79 & 68.83 & 58.44 & 60.09 & 60.01 &54.58 \\
ReLA~\cite{rela} & Swin-B & 63.60 & 62.42 & 56.37 & 70.03 & 69.26 & 59.02 & 61.02 & 59.88 & 58.40 \\
\rowcolor{gray!10} 
\textbf{CoHD} (Ours) & Swin-B & \textbf{68.42} & \textbf{65.17} & \textbf{63.68} & \textbf{72.67} & \textbf{71.85} & \textbf{64.00} & \textbf{63.60} & \textbf{62.63} & \textbf{60.37} \\






\specialrule{.1em}{.05em}{.05em}
\end{tabular}}
}
\end{center}
\vspace{-0.2in}
\caption{Comparison with the state-of-the-art methods on gRefCOCO. 
-V-7B means Vicuna-7B. (ft) denotes that the model is finetuned on the training set of gRefCOCO. $^\dagger$ denotes that the results are from the official repository. $*$ indicates that experimental results using the same hyperparameters configuration from the official repository.}
\vspace{-10pt}
\label{tab:gres}
\end{table*}

%% file: table/rrefcoco.tex
\begin{table*}
   \centering
   \footnotesize
   \renewcommand\arraystretch{1.0}
   \setlength{\tabcolsep}{4.5mm}{
   \resizebox{\linewidth}{!}{
   \begin{tabular}{l|ccc|ccc|ccc}
      \specialrule{.1em}{.05em}{.05em}
      \multirow{2}{*}{Method} & \multicolumn{3}{c|}{R-RefCOCO}  & \multicolumn{3}{c|}{R-RefCOCO+} & \multicolumn{3}{c}{R-RefCOCOg} \\
      \cline{2-10}
                        & mIoU & mRR & rIoU & mIoU & mRR & rIoU & mIoU & mRR & rIoU  \\
      \hline
      CRIS~\cite{cris}  & 43.58 & 76.62 & 29.01 & 32.13 & 72.67 & 21.42 & 27.82	& 74.47 &	14.60 \\
      EFN~\cite{efn}    & 58.33  & 64.64 & 32.53 & 37.74  & 77.12 & 24.24 & 32.53 & 75.33 & 19.44 \\
      VLT~\cite{vlt}    &  61.66  & 63.36 & 34.05 & 50.15 & 75.37 & 34.19 & 49.67 & 67.31 & 31.64\\
      LAVT~\cite{lavt}   & 69.59 & 58.25 & 36.20 & 56.99 & 73.45 & 36.98 & 59.52 &  61.60 & 34.91 \\
      LAVT+~\cite{lavt}     & 54.70 & 82.39 & 40.11 & 45.99 & 86.35 & 39.71 & 47.22 & 81.45 & 35.46 \\
      RefSegformer~\cite{rris} & 68.78 & 73.73 & 46.08 & 55.82 & 81.23 & 42.14 & 54.99 &  71.31 & 37.65 \\ 
      \hline
      \rowcolor{gray!10} 
      \textbf{CoHD} (Ours) & \textbf{74.16} & \textbf{84.27} & \textbf{53.61} & \textbf{64.59} & \textbf{87.49} & \textbf{49.07} & \textbf{63.56} & \textbf{82.68} & \textbf{42.16} \\
      \specialrule{.1em}{.05em}{.05em} 
   \end{tabular}}}
   \vspace{-1.5mm}
   \caption{Comparison with state-of-the-art methods on the R-RefCOCO/+/g dataset.}
   \vspace{-15pt}
   \label{tab:result_R-RIS}
\end{table*}

%% file: sec/4_experiment.tex
\section{Experiments}
\subsection{Experiment Setups}
We evaluate CoHD on three GRES benchmarks using official evaluation metrics: gRefCOCO~\cite{rela}, R-RefCOCO/+/g~\cite{rris} and Ref-ZOM~\cite{refzom}. Following previous GRES methods, we also conduct experiments on the RES dataset: RefCOCO/+/g~\cite{refcoco, refcocog}. Elaborate descriptions of each benchmark and metrics are in the supplementary material. We adopt Swin Transformer~\cite{swin} and BERT~\cite{bert} as the backbone. Due to space limitations, please defer to the supplementary material for complete implementation details.

\input{table/refcoco}
\input{table/ablation_module}

\subsection{Main Results}

\begin{figure*}[t]
    \centering
    \vspace{-10pt}
    \includegraphics[width=0.85\linewidth]{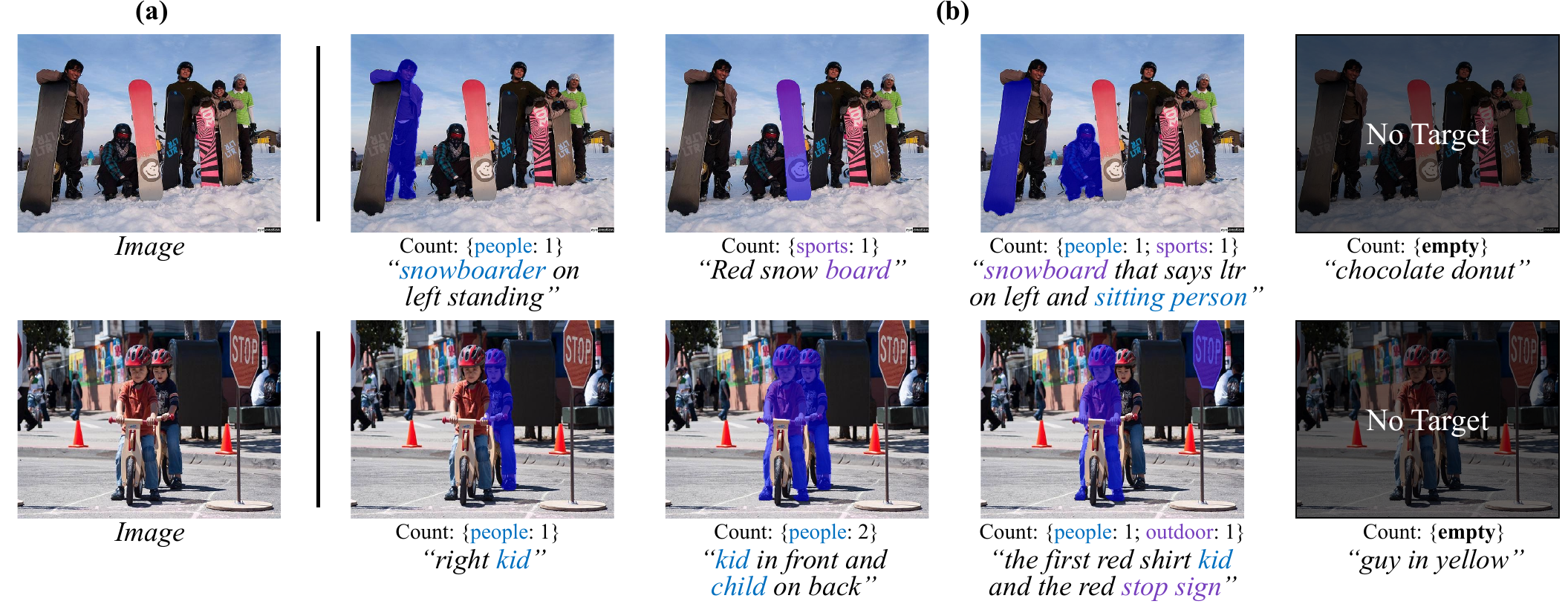}
    \vspace{-10pt}
    \caption{\textbf{Segmentation Results.} (a) and (b) are input images and segmentation results of CoHD with different referring expressions. The term Count specifies the output of AOC module.}
    \label{fig:case_self}
    \vspace{-10pt}
\end{figure*}

\begin{figure*}[t]
    \centering
    \includegraphics[width=0.9\linewidth]{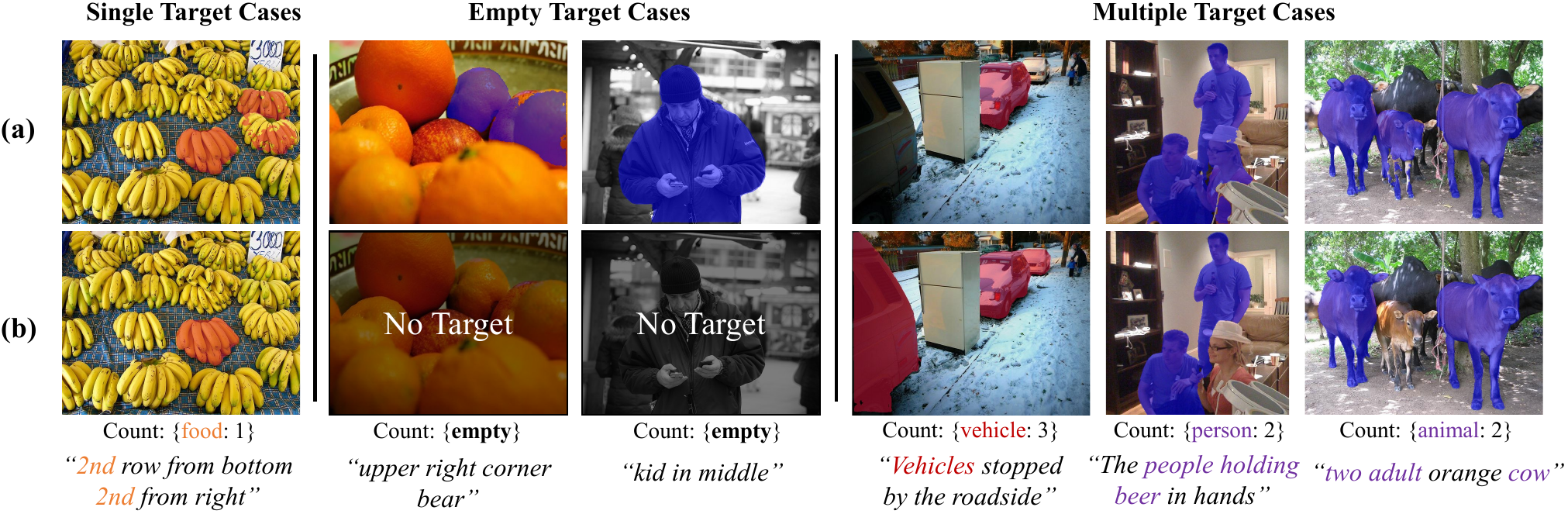}
    \vspace{-10pt}
    \caption{\textbf{Visualization comparison in generalized setting.} (a) and (b) are the segmentation results of ReLA and CoHD, respectively. The term Count specifies the output of AOC module.}
    \label{fig:case}
    \vspace{-15pt}
\end{figure*}

\vspace{-5pt}
\paragraph{Results on GRES tasks.}

To verify the effectiveness of CoHD in the generalized setting, we first compare our method with previous specialist methods in \cref{tab:gres} on gRefCOCO~\cite{rela}. It can be seen that our method achieves new state-of-art performance on all metrics of three evaluation sets in the large-scale GRES benchmark. Under fair comparison with the previous SOTA method ReLA~\cite{rela}, CoHD achieves superior performance with a significant margin of $+4.8\%$, $+2.6\%$, $+2.6\%$ in gIoU of val, testA, testB set. Then, we report our results on R-RefCOCO/+/g~\cite{rris}. As depicted in \cref{tab:result_R-RIS}, our method yields a significant improvement of $+7.5\%$, $+6.9\%$, $+4.5\%$ rIoU in R-RefCOCO/+/g compared with RefSegformer~\cite{rris}.
Due to the similarity of the datasets and space limitations, we report our results on Ref-ZOM benchmark~\cite{refzom} in the appendix. Our CoHD still outperforms all methods even GSVA~\cite{gsva}, a MLLM based model, under a fair setting.

\vspace{-15pt}
\paragraph{Results on RES tasks.}

To fully illustrate the effectiveness of our design, We train CoHD with two settings~\cite{rela, refzom, risclip} in \cref{table:results_refcoco}. CoHD exhibits significant superiority compared with previous methods. With the increment of the semantic context information brought by the data scale, the potential of HSD and AOC can be dramatically exploited.

\subsection{Ablation Study}
In this section, we conduct analysis experiments on the gRefCOCO~\cite{rela} val set using Swin-Tiny as the visual backbone. Note that, we refer to additional ablation studies in the appendix.

\paragraph{Module effectiveness.}
We experiment with the effectiveness of the proposed Hierarchical Semantic Decoder (HSD) and Adaptive Object Counting (AOC) module. As shown in \cref{tab:ablation_modules}, With HSD decouples the intricate referring semantics into the visual-linguistic hierarchy and dynamically aggregates with selective mechanism, the performance is boosted about $4.1\%$ cIoU. With the explicit counting ability, there is an improvement of $2\%$ cIoU. The combination contributes to superior performance on all benchmarks.

\vspace{-5pt}
\paragraph{Analysis of HSD.} 

To investigate the validity of key modules involved in HSD, we provide an analysis of the Semantic Decoding Module (SDM) and Dynamic Hierarchical Aggregation (DHA) components. As \cref{tab:ablation_HSD} shows, the Query Refinement which helps formulate a holistic understanding of the linguistic information with the assistance of the supervised semantic map, brings a clear improvement of $2.5\%$ gIoU. DHA is crucial in excavating the potential of hierarchical benefit, which gains $3.1\%$ gIoU.

\vspace{-5pt}
\paragraph{Variants of object-existence identification.}
We investigate the impact of different variants on the object-existence justification that N-acc can directly reflect.  
As \cref{tab:counting} shows, if maintaining the binary classification across all situations~\cite{rela, rris}, \textit{i.e.}, AOC degraded to an auxiliary task, the counting ability still benefits. We adopt the accuracy between the prediction object count of each category and the true one, termed as C-acc to quantify the counting ability. It can be seen that the simple binary head hampers the object perception, which results in $2.7\%$ C-acc drop. With the subsequent design of counting head upon AOC which facilitates the comprehensive object perception under each situation, CoHD yields the most robust performance.

\paragraph{Analysis of intra- and inter-selection.}
Results on \cref{tab:ablation_DHA} demonstrate that both intra-selection and inter-selection mechanisms are indispensable components in DHA, since the former helps enhance the related region activation at the token level and the latter emphasizes the saliency at the granularity level. The ablation of each will lead to $1.8\%$ and $2.3\%$ degradation respectively.  

\vspace{-10pt}
\paragraph{Mask decoding method.}

To prove the effectiveness of DHA, we compare different prevalent mask decoding methods. As shown in \cref{tab:ablation_decoding}, we observe that the utilization of deep supervision at each level with or without aggregation falls below anticipated outcomes. We speculate that it deteriorates the hierarchical nature that object responses from levels are different but adaptively reciprocal.

\vspace{-10pt}
\paragraph{Comprehensive evaluation in generalized settings.}

To quantify our overall superiority in addressing the challenges introduced in GRES, we synthetically compare CoHD with the previous SOTA method, ReLA~\cite{rela} from three perspectives: 1) gIoU, obtained from only multiple-target scenarios. 2) N-acc., calculated in non-target cases individually. 3) Pr@0.7, measured by the single-target situation. As illustrated in \cref{tab:comprehensive_evalution}, CoHD shows absolute dominance across dimensions compared to ReLA.   

\subsection{Qualitative Results}
Benefiting from the dynamic hierarchical aggregation on well-aligned semantics of different granularity and the counting ability, \cref{fig:case_self} and \cref{fig:case} demonstrate our superiority in processing the complex spatial relationship between instances and intricate referring text in multiple/single/non-target scenarios. In contrast, ReLA~\cite{rela} fails to adequately address these challenges in GRES due to its 
incomplete multi-granularity object information modeling.
More visualization results are in the supplementary material.

%% file: table/refcoco.tex
\begin{table*}
    \centering
     \footnotesize
    \vspace{-3.6mm}
     \setlength{\tabcolsep}{4.0mm}
     {\begin{tabular}{l|c|ccc|ccc|cc}
      \specialrule{.1em}{.05em}{.05em} 
        \multirow{2}{*}{Methods} &\multirow{2}{*}{Backbone} & \multicolumn{3}{c|}{RefCOCO} & \multicolumn{3}{c|}{RefCOCO+} & \multicolumn{2}{c}{RefCOCOg} \\
        \cline{3-10}
          & & val   & test A & test B & val   & test A & test B & val$_\text{(U)}$   & test$_\text{(U)}$  \\
        \midrule
        \multicolumn{10}{c}{\textit{Single Dataset} (oIoU)} \\
        \midrule
        MCN~\cite{mcn} &Darknet53& 62.44 & 64.20 & 59.71 & 50.62 & 54.99 & 44.69 & 49.22 & 49.40      \\
        {VLT} \cite{vlt} &Darknet53 & 67.52 & 70.47 & 65.24 & 56.30 & 60.98 & 50.08 & 54.96 & 57.73  \\
        {ReSTR}~\cite{Restr} &ViT-B& 67.22 & 69.30 & 64.45 & 55.78 & 60.44 & 48.27 & -     & -     \\
        {CRIS}~\cite{cris}& CLIP-R101& 70.47 & 73.18 & 66.10 & 62.27 & 68.08 & 53.68 & 59.87 & 60.36   \\
        {LAVT}~\cite{lavt}&{Swin-B}& 72.73 & {75.82} & 68.79 & 62.14 & 68.38 & 55.10 & 61.24 & 62.09  \\
        RefSegformer~\cite{rris} & {Swin-B} & 73.73 & 75.64 & 70.09 & 63.50 & 68.69 & 55.44 & 62.56 & 63.07 \\
        RISCLIP~\cite{risclip} & CLIP-B &  73.57 & 76.46 & 69.76 & 65.53 & 70.61 & 55.49 & 64.10 & 65.09 \\
        {ReLA}~\cite{rela} &{Swin-B} &  73.82 & 76.48 & 70.18 & 66.04 & 71.02 & 57.65 & 65.00 & 65.97 \\
        {DMMI}~\cite{refzom} &{Swin-B} & 74.13 & 77.13 & 70.16 & 63.98 & 69.73 & 57.03 & 63.46     & 64.19     \\
        \midrule
        \rowcolor{gray!10} \textbf{CoHD} &{Swin-B}& \textbf{75.12} & \textbf{78.33} & \textbf{71.00} & \textbf{66.79} & \textbf{71.58} & \textbf{58.37} & \textbf{65.84}     & \textbf{66.70}    \\
        \midrule
        \multicolumn{10}{c}{\textit{Combined Dataset} (mIoU)}\\
        \midrule
        PolyFormer$^{\dagger}$~\cite{polyformer} & Swin-B & 75.96 & 77.09 & 73.22 & 70.65 & 74.51 & 64.64 & 69.36 & 69.88 \\
        RISCLIP~\cite{risclip} & CLIP-B &  76.01 & 78.63 & 71.94 & 69.67 & 74.30 & 61.37 & 69.61 & 69.56 \\
        \rowcolor{gray!10} \textbf{CoHD} &{Swin-B}& \textbf{78.11} & \textbf{80.39} & \textbf{75.20} & \textbf{72.03} & \textbf{76.37} & \textbf{65.45} & \textbf{70.83}     & \textbf{72.11}    \\
        \specialrule{.1em}{.05em}{.05em} 
     \end{tabular}}
    \vspace{-0.1in}
    \caption{Comparison with previous state-of-the-art GRES methods on classic RES benchmarks under different settings. $^{\dagger}$ indicates that the training includes additional datasets for pretraining.}
    \label{table:results_refcoco}
\end{table*}

%% file: table/ablation_module.tex
\begin{table*}[t]
\vspace{-6pt}
\begin{minipage}[c]{\textwidth}
    \begin{minipage}{0.28\textwidth}
    \makeatletter\def\@captype{table}
    \centering
    \footnotesize
    \setlength{\tabcolsep}{3.0pt}
    \begin{tabular}{c c | c c c}
    \toprule
    HSD &  AOC & gIoU & cIoU & N-acc.\\
    \midrule
      &   & 53.52 & 56.55 & 41.11 \\
      &  \checkmark  & 57.63 & 58.56 & 49.99 \\
    \checkmark &    &  63.15 & 60.67 & 55.84 \\
    \checkmark & \checkmark & \textbf{65.89} & \textbf{62.95} & \textbf{60.95} \\
    \bottomrule
    \end{tabular}
    \caption{Effectiveness of the modules.}
    \label{tab:ablation_modules}
    \end{minipage}
    \begin{minipage}{0.37\textwidth} %
    \makeatletter\def\@captype{table}
    \centering
    \footnotesize
    \setlength{\tabcolsep}{2.8pt}
    \begin{tabular}{cc|ccc}
    \toprule
    Intra-Select & Inter-Select & gIoU & cIoU & N-acc.\\
    \midrule
     &  & 62.77 & 62.48 & 53.78\\
     \checkmark &  & 63.56 & 62.32 & 56.16 \\
      & \checkmark & 64.03 & 62.39 & 56.18 \\
    \checkmark & \checkmark & \textbf{65.89} & \textbf{62.95} & \textbf{60.95} \\
    \bottomrule
    \end{tabular}
    \caption{Effects of components of DHA module.}
    \label{tab:ablation_DHA}
    \end{minipage}
    \begin{minipage}{0.35\textwidth} 
    \makeatletter\def\@captype{table}
    \centering
    \footnotesize
    \setlength{\tabcolsep}{2.0pt} 
    \begin{tabular}{cc|ccc}
    \toprule
        Method & Backbone & gIoU & N-acc. & Pr@0.7 \\
    \midrule
        ReLA~\cite{rela} & Swin-T & 70.30 & 44.07 & 52.21 \\
        ReLA~\cite{rela} & Swin-B & 73.47 & 56.37 & 59.89 \\
        CoHD & Swin-T & 73.07 & 60.95 & 62.62 \\
        CoHD & Swin-B & \textbf{75.78} & \textbf{63.68} & \textbf{66.24} \\
    \bottomrule
    \end{tabular}
    \caption{Comprehensive evaluation.}
    \label{tab:comprehensive_evalution}
    \end{minipage}
    \vfill
    \begin{minipage}{0.4\textwidth}
    \makeatletter\def\@captype{table}
    \centering
    \footnotesize
    \vspace{5pt}
    \setlength{\tabcolsep}{2.0pt}
    \begin{tabular}{cc|cccc}
    \toprule
        AOC & Object Indicator & gIoU & cIoU & N-acc. &  C-acc. \\
        \midrule
         & binary head & 63.15 & 60.67 & 55.84 &  --  \\ 
        \checkmark &  binary head & 63.72  & 61.59  & 56.09  & 72.73 \\ 
        \checkmark &  count head & \textbf{65.89} & \textbf{62.95} & \textbf{60.95} & \textbf{75.45} \\
        \bottomrule
    \end{tabular}
    \caption{Impact of different strategies of object-existence identification.}
    \label{tab:counting}
    \end{minipage}
    \begin{minipage}{0.35\textwidth}
    \makeatletter\def\@captype{table}
    \centering
    \footnotesize
    \vspace{-5pt}
    \setlength{\tabcolsep}{2.0pt}
    \begin{tabular}{ccc|ccc}
    \toprule
       Mask Gen. & Query Ref. & DHA & gIoU & cIoU & N-acc. \\
        \midrule
        \checkmark& & \checkmark & 63.31  & 62.74  & 54.15  \\
        \checkmark& \checkmark & & 62.77 & 62.48 & 53.78 \\ 
        \checkmark& \checkmark & \checkmark & \textbf{65.89} & \textbf{62.95} & \textbf{60.95} \\ 
    \bottomrule
    \end{tabular}
    \caption{Impact of different structure of HSD.}
    \label{tab:ablation_HSD}
    \end{minipage}
    \begin{minipage}{0.23\textwidth}
    \makeatletter\def\@captype{table}
    \centering
    \footnotesize
    \vspace{5pt}
    \setlength{\tabcolsep}{4.5pt}
    \begin{tabular}{c|cc}
    \toprule
        Mask Decode & gIoU & cIoU \\ \midrule
        Deep Sup. & 64.11  & 61.98 \\
        DHA  & \textbf{65.89} & \textbf{62.95} \\ 
        Both & 63.35 & 62.34 \\
    \bottomrule
    \end{tabular}
    \caption{Comparison of decoding methods.}
    \label{tab:ablation_decoding}
    \end{minipage}
\end{minipage}
\vspace{-15pt}
\end{table*}

%% file: sec/5_conclusion.tex
\section{Conclusion}
This paper proposes a framework called CoHD for GRES, aiming to decouple the complex referring semantics into different levels for boosting multi-granularity comprehension. With the well-aligned visual-linguistic hierarchy of different granularity, CoHD dynamically aggregates it with the intra- and inter-selection. Moreover, with complete semantic context, we empower CoHD with category-specific counting ability to promote the overall object perception in multiple/single/non-target scenarios. 
Experimental results on GRES benchmarks show the effectiveness of CoHD and the superiority in addressing the new challenges.

%% file: sec/appendix.tex
\clearpage
\section*{Appendix}
\section{Additional Details on Experiment Setup} 
\subsection{Datasets}
\label{appendix:dataset}

\paragraph{gRefCOCO.} It contains 278,232 expressions, which includes 80,022 multiple-target referents and 32,202 empty-target ones. There are 60,287 distinct instances being referred in 19,994 images. The images are split into four subsets: training, validation, test-A, and test-B following the same UNC partition of RefCOCO~\cite{refcoco}.
\paragraph{Ref-ZOM.} Ref-ZOM are selected from COCO dataset~\cite{coco}, which consists of 55,078 images and 74,942 annotated objects. 43,749 images and 58,356 objects are utilized in training, and 11,329 images and 16,586 objects are employed in testing. It is annotated with three different settings, \textit{i.e.}, one-to-zero, one-to-one, one-to-many, each of which corresponds to the empty-target, single-target, and multiple-target in GRES respectively.
\paragraph{R-RefCOCO.} There are three different sets in the dataset, R-RefCOCO, R-RefCOCO+, R-RefCOCOg, and each of which is based on the classic RES benchmark, RefCOCO+/g~\cite{refcoco}. Only the validation set follows the UNC partition principle and it is officially stated for evaluation. The formulation rule of the dataset is adding negative sentences into the training set at a 1:1 ratio relative to the positive sentences.
\paragraph{RES.} RefCOCO~\cite{refcoco}, RefCOCO+~\cite{refcoco}, and RefCOCOg~\cite{refcocog} are three standard RES benchmarks, each of which contains 19,994, 19,992, and 26,711 images, with
50,000, 49,856, and 54,822 annotated objects and 142,209,
141,564, and 104,560 annotated expressions, respectively. 

\subsection{Metrics} 
\label{appendix:metrics}
For GRES, following \cite{rela}, we measure the effectiveness of our model by Pr@0.7, gIoU, cIoU and N-acc for gRefCOCO. Meantime, oIoU, mIoU, Acc. used in~\cite{refzom} are adopted for Ref-ZOM. In addition, the standard metrics mIoU, mRR, rIoU are for R-RefCOCO~\cite{rris}. It is important to note that these metrics are officially specified in each respective benchmark. 

Similar to the mean IoU, the Generalized IoU (gIoU) computes the average IoU value for each image across all instances. For the empty-target cases, the IoU values for true positive empty-target instances are considered to be $1$ whereas the IoU values for false negative instances are deemed to be $0$. cIoU calculates the total intersection pixels over the total union pixels. 

mIoU, oIoU is adopted in Ref-ZOM~\cite{refzom}, where mIoU is the average IoU value for each image across all cases containing referred objects. oIoU is the same as cIoU. As for R-RefCOCO~\cite{rris}, we use the metric rIoU for quantifying the quality of robust segmentation, which takes the negative sentences into consideration and explicitly assigns the equal weight of the positive one in mIoU calculation. 
Note that, N-acc. in gRefCOCO, Acc. in Ref-ZOM are in the same formulation where they denote the ratio of the correctly classified empty-target expressions over all the empty-target expressions in the dataset. Similarly, mRR in R-RefCOCO calculates the empty-target expression recognition rate of each image and averages these across the entire dataset.

\subsection{Implementation Details} \label{appendix:implementation}
\paragraph{Experiment setup.}
Our model is implemented with detectron2~\cite{detectron2} in Pytorch. The visual encoder is initialized with the pre-trained weights on ImageNet~\cite{imagenet} and the language encoder is an officially pre-trained BERT model~\cite{bert}. We set the number of Deformable attention layers as $6$ following ReLA~\cite{rela}. There are $3$ cascaded semantic decoding modules in HSD for mask generation and query refinement. The weight of $\mathit{Loss}_{mask}$ and $\mathit{Loss}_{count}$ are set as $2$ and $0.1$ by default. It is worth noting that since $\mathit{Loss}_{exist}$ is trained individually, which has no impact on the main framework, we directly set its weight as $1$ for all experiments. 

The model is trained with AdamW optimizer with a weight decay of 0.05. The batch size is set to 48. The learning rate is initialed as 2e-5 and scheduled by cosine learning rate decay by default. Following \cite{lavt, rela}, the input images are resized to 480$\times$ 480 and the maximum length of referring expressions is set as $20$ for all datasets. Other hyperparameters of the encoding process are the same as ReLA~\cite{rela}. All experiments are conducted with $8\times$ A10 and each takes up about one day with 13GB $\sim$ 18GB memory occupied \textit{i.e.}, it depends on the backbone.

\paragraph{Counting labels formulation.} We elaborate on the formulation of the counting label. All mentioned datasets adhere to the COCO ~\cite{coco} annotation format. On the one hand, for RES datasets, each annotated expression is accompanied by a label (\textit{category\_id}) corresponding to the target category. On the other hand, a list of target categories is incorporated with the additional multi-target scenario for GRES dataset. Consequently, the count of objects can be derived from the number of categories and each classification label is from the given category of the annotated object. It signifies that the construction of counting ground truth $\mathcal{C}^{gt}$ is straightforward and the additional information extraction is unnecessary. Note that, taking the long-tail distribution of the 80 original COCO categories into consideration where most of the categories are annotated as $0$, we instead utilize the 12 super-categories to narrow down the referential search space for providing more precise supervision.

\input{table/refzom}

\input{table/ablation_parameters}

\section{Additional Experiments}

\subsection{Discussion on Inconsistent Performance} \label{appendix:res_result}
As shown in the manuscript, we notice that the performance improvement is inconsistent between RES and GRES. We believe it can be attributed to two folds: 1) \textbf{\textit{Restricted effectiveness on the generalized design}}: It is obvious that the diversity of referring scenarios in GRES is much more plentiful compared to the RES since RES only includes one-to-one referent case. Due to the inadequate referring semantics between visual and linguistic and the lack of enriched contextual information, \textit{e.g.}, Spatial relationship between instances, counting, or compound structure expressions, it is believed that the great potential of HSD is constrained. Moreover, the effectiveness of object counting mechanism in RES is also underestimated. In GRES, it seamlessly integrates all specificities of each scenario by embodying each case into count- and category-level supervision. In terms of RES, the simplification of the formulation limits the potential of object counting, where count number supervision is missing. 2) \textbf{\textit{Imbalance dataset scale}}: The samples of GRES dataset gRefCOCO is 230,944 samples, while that of each RES dataset is: RefCOCO: 120,624 RefCOCO+: 120,191 RefCOCOg: 80,544. That indicates the specific design and effort on hyper-parameter studies in RES weighs more crucial. Since our main focus is on the generalized referring which accompanies more useful real-world applications, complex hyperparameter studies on RES are not applied. 

Considering the hypothesis mentioned above, we believe that enlarging the scale of the dataset can partly alleviate the phenomenon. The results in the joint dataset training~\cite{risclip} demonstrate that our CoHD brings more improvements compared with the single scenario, showing that the potential of our generalized paradigms can be exploited with enriched contextual information. 

\subsection{Performance on Ref-ZOM Dataset}
We report our results on Ref-ZOM benchmark~\cite{refzom} in \cref{table:ref_zom}. As illustrated, our method outperforms all methods under a fair setting, \textit{e.g.}, $+6.3\%$ in Acc., $+1.6\%$ in mIoU. It is worth noting that our method is better than GSVA~\cite{gsva}, which utilizes Multi-Modal Large Model (MLLM)~\cite{llava}. 
\input{table/ablation_num_parameters}
\input{table/ablation_stacked_layer}

\subsection{Additional Ablation Studies}
\label{appendix:ablation_studies}

\paragraph{Loss ratio.}
$\lambda_{mask}$ and $\lambda_{count}$ are the coefficients for $\mathit{Loss}_{mask}$ and $\mathit{Loss}_{count}$ respectively. We report the ablation results in \cref{tab:ablation_loss}. As observed, the appropriate settings of $\lambda_{mask}$ and $\lambda_{count}$ help decently integrate the counting ability into hierarchical semantic decoding. 

\paragraph{Stacked layer in Semantic Decoding Module.}
We experiment with the impact on the number of stacked layers in the semantic decoding module. As shown in \cref{tab:ablation_layer}, the insufficient or excessive layers both lead to decreased performance due to incomplete semantic context modeling or over-exaggeration of the redundant contents.

\paragraph{Efficiency \textit{v.s.} performance.}
We provide detailed comparisons (including model parameters and T-FLOPs.) on the gRefCOCO val set and the results are shown in \cref{tab:ablation_performance_efficiency}. It can be seen that our CoHD-B outperforms previous SOTA methods ReLA-B by 4.8\% gIoU and 2.7\% cIoU with slight parameters increased, \textit{i.e.}, 22M, and even no T-FLOPs costs. It is worth noting that the performance of CoHD-T is even better than ReLA-B with fewer parameters. In addition, the number of parameters consumed by HSD is 14.7\% of the whole model, but with significant performance improvement.


\section{Visualizations.} \label{appendix:visualization}

\subsection{Multi-granularity Mask Aggregation Visualization}
\begin{figure*}
    \centering
    \includegraphics[width=\linewidth]{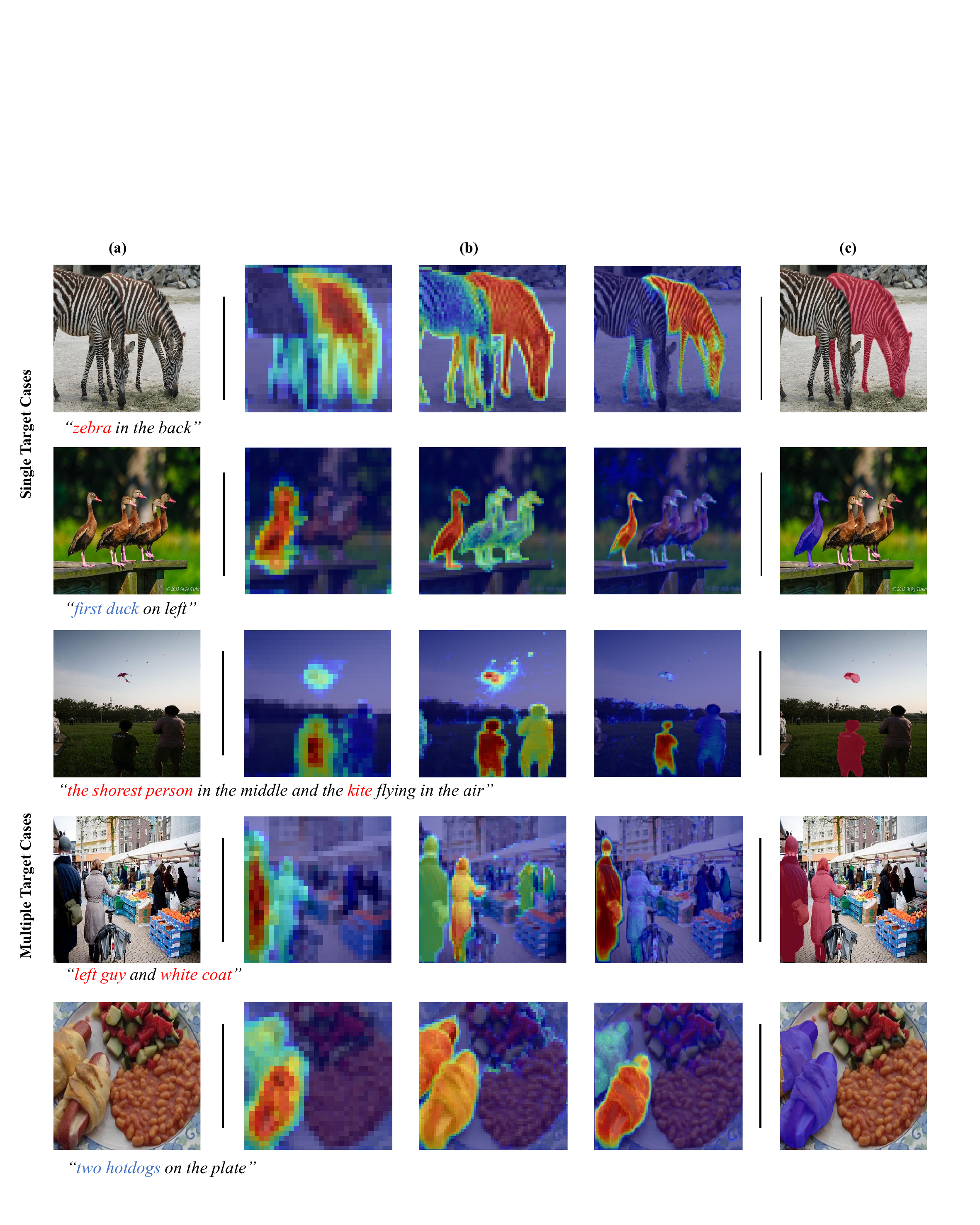}
    \vspace{-15pt}
    \caption{\textbf{Visualization in Multi-granularity Mask Aggregation.} (a) and (c) indicate the input image and corresponding segmentation result of our CoHD, respectively. (b) illustrates the aggregated activation map at each granularity.
}
    \label{fig:attn_map}
    \vspace{-10pt}
\end{figure*}

As illustrated in the main paper, we hierarchically aggregate each visual-linguistic correspondence in different granularity for hierarchical semantic decoding. To further prove the rationality of our design, we visualize the aggregated activation map (originally from the semantic map) at each level in \cref{fig:attn_map}. As observed, with progressive mutual modality complementary across granularities, the desired regions of the referent target are fully activated with the increase of the levels.

\subsection{Additional Segmentation Comparisons}
\begin{figure*}
    \centering
    \includegraphics[width=0.97\linewidth]{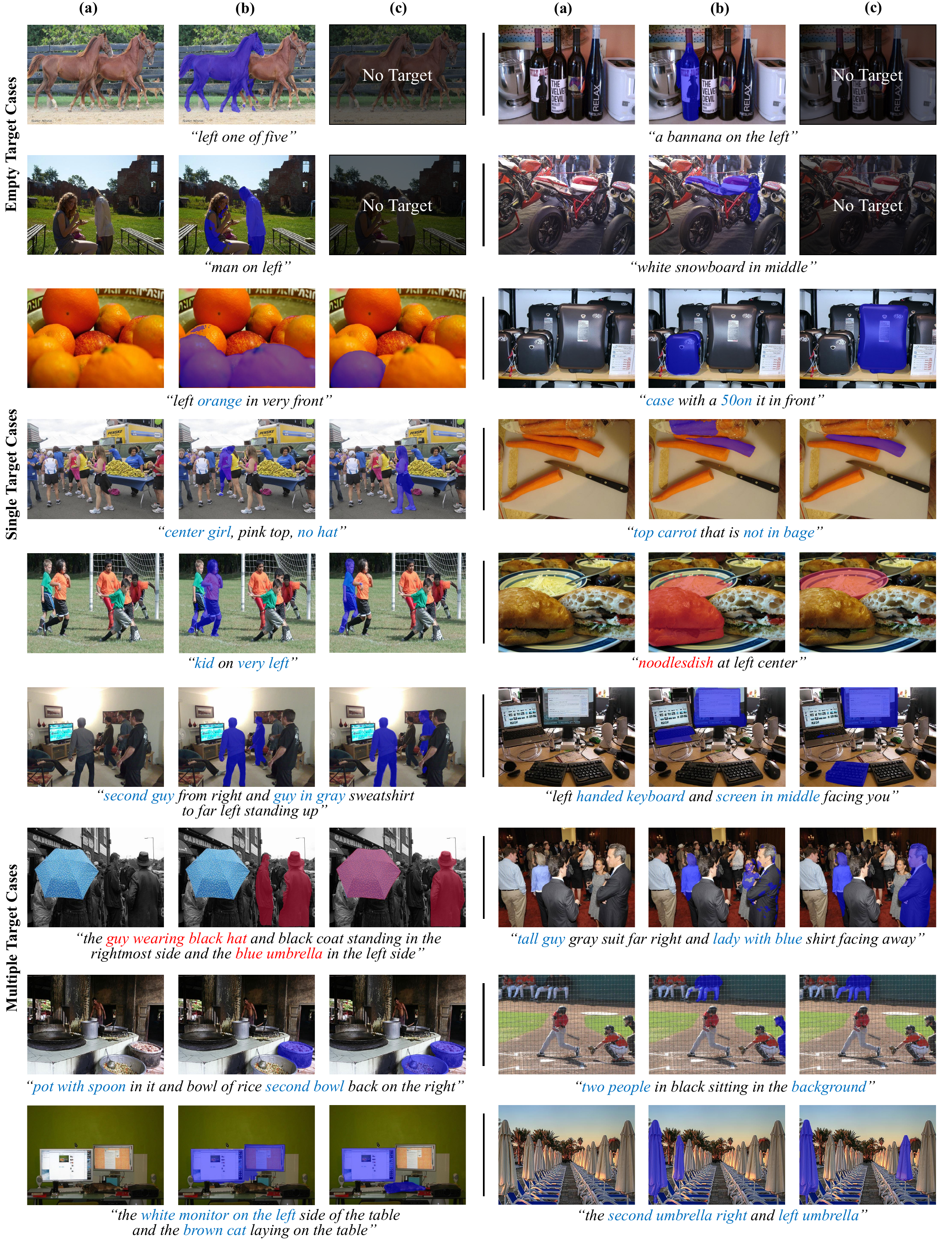}
    \vspace{-10pt}
    \caption{\textbf{Segmentation results comparison in the generalized setting.} (a) denotes the input image, (b) and (c) are segmentation results of ReLA and CoHD, respectively. Empty, Single, and Multiple target referent situations with different referring expressions are in top-to-bottom accordance.
}
    \label{fig:case4}
\end{figure*}

As illustrated in the main paper, CoHD can better handle the extreme challenges of GRES under multiple/single/empty target scenarios compared with the previous SOTA method ReLA~\cite{rela}. Here, we incorporate more cases in \cref{fig:case4} to demonstrate our superiority in modeling the intricate referring association between visual and linguistic. 

\subsection{Segmentation Results of CoHD}
\begin{figure*}
    \centering
    \includegraphics[width=\linewidth]{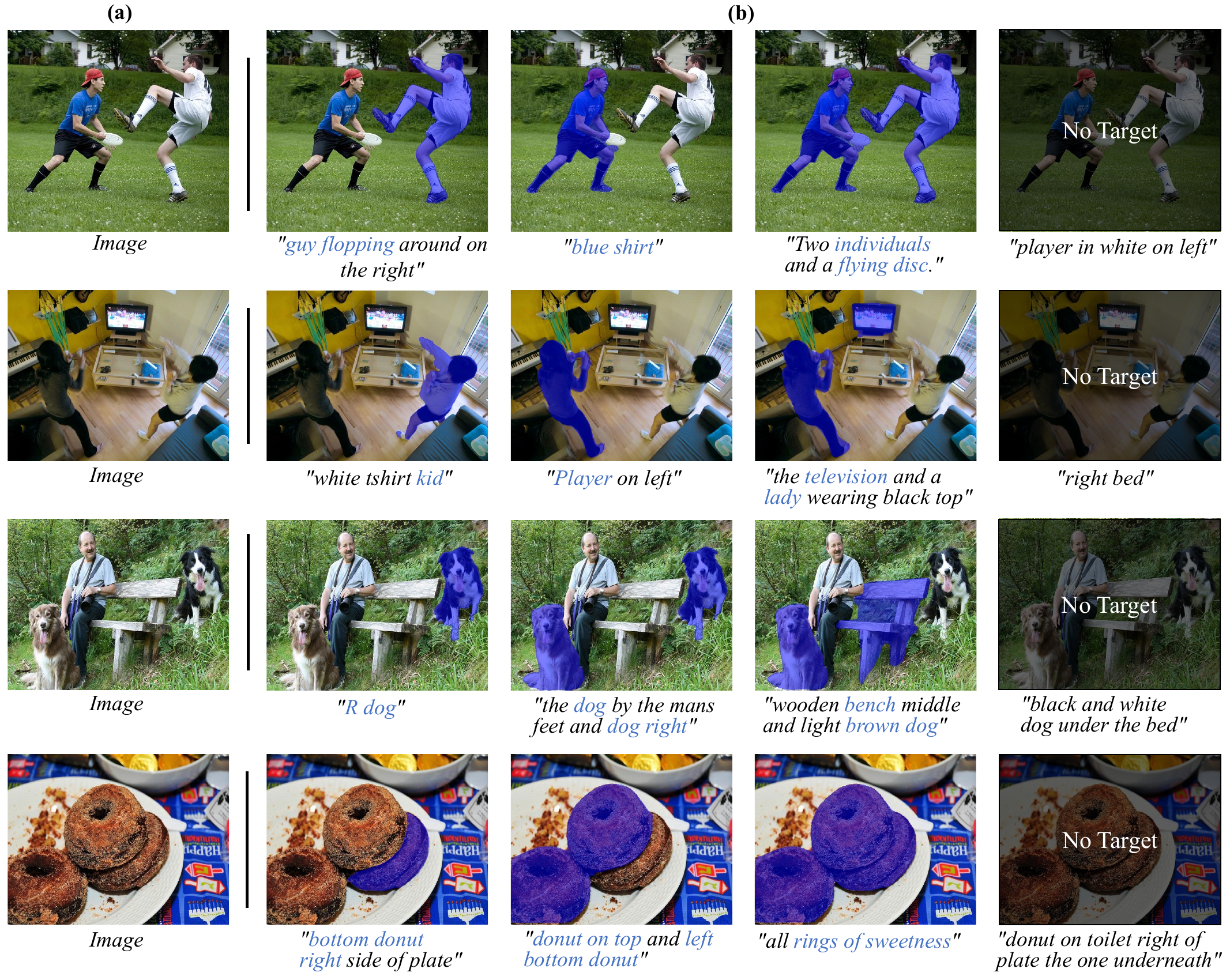}
    \vspace{-15pt}
    \caption{\textbf{Segmentation results of CoHD in generalized settings.} (a) denotes the input image, and (b) showcases the segmentation results of CoHD under multiple/single/non-target situations with different referring expressions.
}
    \label{fig:case_more}
    \vspace{-10pt}
\end{figure*}

\cref{fig:case_more} demonstrates the effectiveness of CoHD in complex generalized settings, \textit{e.g.}, complex geometrical relationships between instances, deceptive objects, compound sentence structures, and intricate associations between referring expressions and images.

\section{Limitations} \label{appendix:limitation}
Benefiting from the more compatible design, \textit{i.e.}, hierarchical semantic decoding paradigm and the explicit counting ability, our method CoHD sufficiently addresses the limitations of existing GRES and achieves superiority in meeting the challenges of GRES. However, there are some potential limitations. Since the referring sentences in GRES contain multiple-target expressions, the truncation of the input text into $20$ may lose detailed descriptions of some of the targets. Although we compensate it with sentence-level textual features, it still remains unsatisfactory in terms of the incompleteness of fine-grained target description. We believe that how to fully utilize the textual expression in GRES is an interesting future direction. 

\clearpage
\twocolumn

%% file: table/refzom.tex
\begin{table}
\begin{center}
\setlength{\tabcolsep}{2mm}{
\renewcommand\arraystretch{1.0}
\resizebox{\linewidth}{!}{
\begin{tabular}{l|c|ccc}
\specialrule{.1em}{.05em}{.05em}
\multirow{2}{*}{Method}& \multirow{2}{*}{Backbone} & \multicolumn{3}{c}{Ref-ZOM Test Set} \\
 & & mIoU & oIoU & Acc. \\ 
\midrule
\multicolumn{5}{c}{\textit{MLLM Methods}} \\
\midrule
LISA-V-7B~\cite{lisa}& SAM-ViT-H & 61.46 & 60.14  & 72.58 \\

GSVA-V-7B~\cite{gsva} & SAM-ViT-H & 67.98 & 67.12  & 82.66 \\
LISA-V-7B~\cite{lisa} (ft) & SAM-ViT-H & 65.39 & 66.41  & 93.39 \\
GSVA-V-7B~\cite{gsva} (ft) & SAM-ViT-H & 68.13 & 68.29  & 94.59 \\
\midrule
\multicolumn{5}{c}{\textit{Specialist Methods}} \\
\midrule
MCN~\cite{mcn}   & DarkNet-53   & 54.70  & 55.03   & 75.81        \\ 
CMPC~\cite{cmpc}  & ResNet-101    & 55.72   & 56.19    & 77.01        \\ 
VLT~\cite{vlt}   & DarkNet-53     & 60.43    & 60.21    & 79.26        \\ 
LAVT~\cite{lavt}  & Swin-B     & 64.78   & 64.45    & 83.11        \\ 
DMMI~\cite{refzom}  & Swin-B   & 68.21   & 68.77   & 87.02  \\
\midrule

\rowcolor{gray!10} \textbf{CoHD} (Ours) & Swin-B & \textbf{69.81} & \textbf{68.99} & \textbf{93.34}\\

\specialrule{.1em}{.05em}{.05em}
\end{tabular}}}
\vspace{1mm}
\caption{Comparison with state-of-the-art methods on the Ref-ZOM dataset.}
\label{table:ref_zom}
\vspace{-0.1in}
\end{center}
\end{table}

%% file: table/ablation_parameters.tex

\begin{table}
\begin{center}
\setlength{\tabcolsep}{3.0mm}{
\renewcommand\arraystretch{1.0}
\resizebox{\linewidth}{!}{
\begin{tabular}{cc|cccc}
    \toprule
    $\lambda_{mask}$ & $\lambda_{count}$ & gIoU & cIoU & N-acc. & C-acc.\\
    \midrule
     $2$ & $0.5$ & 63.23 & 62.37 & 54.88 & 73.28 \\
     $1$ & $0.5$ & 62.39 & 61.53 & 54.81 & 73.20 \\
     $1$ & $1$ & 63.06 & 61.80 & 55.34 & 71.88 \\
     $5$ & $0.1$ & 64.15 &  62.54 & 57.24 & 71.46 \\
     $2$ & $0.1$ & \textbf{65.89} & \textbf{62.95} & \textbf{60.95} & \textbf{75.45}\\
    \bottomrule
    \end{tabular}}
    \caption{Results of different ratios of loss.}
    \label{tab:ablation_loss}
\vspace{-15pt}}
\end{center}
\end{table}

%% file: table/ablation_num_parameters.tex
\begin{table}
    \centering
    \setlength{\tabcolsep}{1.0mm}{
    \renewcommand\arraystretch{1.0}
    \resizebox{\linewidth}{!}{
    \begin{tabular}{ccccccc}
    \toprule
    Method & Backbone & Parameters & T-FLOPs & gIoU & cIoU & N-acc. \\
    \midrule
    ReLA & Swin-T & 163M & 0.066T & 56.87&  57.73 & 44.07 \\
    CoHD & Swin-T & 185M  & 0.068T & 65.89 & 62.95 & 60.95 \\
    \midrule
    DMMI & Swin-B & 341M & 0.392T & 62.68 & 62.77 & 53.20 \\
    ReLA & Swin-B & 226M & 0.131T & 63.60 & 62.42 & 56.37 \\
    CoHD & Swin-B & 248M  & 0.133T & 68.42 & 65.17 & 63.68 \\
    \bottomrule
    \end{tabular}}}
    \caption{Performance and efficiency comparison with previous SOTA method under different backbones.}
    \label{tab:ablation_performance_efficiency}
\end{table}

%% file: table/ablation_stacked_layer.tex
\begin{table}
    \centering
    \setlength{\tabcolsep}{3.0mm}{
    \renewcommand\arraystretch{1.0}
    \resizebox{0.8\linewidth}{!}{
    \begin{tabular}{cccc}
    \toprule
    Stacked Layer & gIoU & cIoU & N-acc. \\
    \midrule
    2 & 66.33 & 64.58 & 58.48 \\
    3 & 68.42 & 65.17 & 63.68 \\
    4 & 66.70 &  64.94 & 60.49\\
    \bottomrule
    \end{tabular}}}
    \caption{Impact on different numbers of stacked layers in the Semantic Decoding Module.}
    \label{tab:ablation_layer}
\end{table}

%% file: main.bbl
\begin{thebibliography}{38}
\providecommand{\natexlab}[1]{#1}
\providecommand{\url}[1]{\texttt{#1}}
\expandafter\ifx\csname urlstyle\endcsname\relax
  \providecommand{\doi}[1]{doi: #1}\else
  \providecommand{\doi}{doi: \begingroup \urlstyle{rm}\Url}\fi

\bibitem[Chen et~al.(2019)Chen, Jia, Lo, Chen, and Liu]{chensee2019}
Ding{-}Jie Chen, Songhao Jia, Yi{-}Chen Lo, Hwann{-}Tzong Chen, and Tyng{-}Luh Liu.
\newblock See-through-text grouping for referring image segmentation.
\newblock In \emph{ICCV}, pages 7453--7462. {IEEE}, 2019.

\bibitem[Cholakkal et~al.(2019)Cholakkal, Sun, Khan, and Shao]{cholakkal2019}
Hisham Cholakkal, Guolei Sun, Fahad~Shahbaz Khan, and Ling Shao.
\newblock Object counting and instance segmentation with image-level supervision.
\newblock In \emph{CVPR}, pages 12397--12405, 2019.

\bibitem[Deng et~al.(2009)Deng, Dong, Socher, Li, Li, and Fei-Fei]{imagenet}
Jia Deng, Wei Dong, Richard Socher, Li-Jia Li, Kai Li, and Li Fei-Fei.
\newblock Imagenet: A large-scale hierarchical image database.
\newblock In \emph{2009 IEEE conference on computer vision and pattern recognition}, pages 248--255. Ieee, 2009.

\bibitem[Devlin et~al.(2019)Devlin, Chang, Lee, and Toutanova]{bert}
Jacob Devlin, Ming{-}Wei Chang, Kenton Lee, and Kristina Toutanova.
\newblock {BERT:} pre-training of deep bidirectional transformers for language understanding.
\newblock In \emph{NAACL}, pages 4171--4186, 2019.

\bibitem[Ding et~al.(2021)Ding, Liu, Wang, and Jiang]{vlt}
Henghui Ding, Chang Liu, Suchen Wang, and Xudong Jiang.
\newblock Vision-language transformer and query generation for referring segmentation.
\newblock In \emph{ICCV}, pages 16301--16310, 2021.

\bibitem[Feng et~al.(2021)Feng, Hu, Zhang, and Lu]{efn}
Guang Feng, Zhiwei Hu, Lihe Zhang, and Huchuan Lu.
\newblock Encoder fusion network with co-attention embedding for referring image segmentation.
\newblock In \emph{CVPR}, pages 15506--15515. Computer Vision Foundation / {IEEE}, 2021.

\bibitem[Goldman et~al.(2019)Goldman, Herzig, Eisenschtat, Goldberger, and Hassner]{goldman2019}
Eran Goldman, Roei Herzig, Aviv Eisenschtat, Jacob Goldberger, and Tal Hassner.
\newblock Precise detection in densely packed scenes.
\newblock In \emph{CVPR}, pages 5227--5236, 2019.

\bibitem[Hsieh et~al.(2017)Hsieh, Lin, and Hsu]{heish2017}
Meng{-}Ru Hsieh, Yen{-}Liang Lin, and Winston~H. Hsu.
\newblock Drone-based object counting by spatially regularized regional proposal network.
\newblock In \emph{ICCV}, pages 4165--4173, 2017.

\bibitem[Hu et~al.(2016)Hu, Rohrbach, and Darrell]{husegmentation2016}
Ronghang Hu, Marcus Rohrbach, and Trevor Darrell.
\newblock Segmentation from natural language expressions.
\newblock In \emph{ECCV}, pages 108--124, 2016.

\bibitem[Hu et~al.(2023)Hu, Wang, Shao, Xie, Li, Han, and Luo]{refzom}
Yutao Hu, Qixiong Wang, Wenqi Shao, Enze Xie, Zhenguo Li, Jungong Han, and Ping Luo.
\newblock Beyond one-to-one: Rethinking the referring image segmentation.
\newblock In \emph{ICCV}, pages 4044--4054, 2023.

\bibitem[Huang et~al.(2020)Huang, Hui, Liu, Li, Wei, Han, Liu, and Li]{cmpc}
Shaofei Huang, Tianrui Hui, Si Liu, Guanbin Li, Yunchao Wei, Jizhong Han, Luoqi Liu, and Bo Li.
\newblock Referring image segmentation via cross-modal progressive comprehension.
\newblock In \emph{CVPR}, pages 10485--10494, 2020.

\bibitem[Huang et~al.(2023)Huang, Dai, Zhang, Zhang, and Shan]{psc}
Zhizhong Huang, Mingliang Dai, Yi Zhang, Junping Zhang, and Hongming Shan.
\newblock Point, segment and count: {A} generalized framework for object counting.
\newblock \emph{arXiv preprint arXiv:2311.12386}, 2023.

\bibitem[Jiang et~al.(2024)Jiang, Sablayrolles, Roux, Mensch, Savary, Bamford, Chaplot, Casas, Hanna, Bressand, et~al.]{moe}
Albert~Q Jiang, Alexandre Sablayrolles, Antoine Roux, Arthur Mensch, Blanche Savary, Chris Bamford, Devendra~Singh Chaplot, Diego de~las Casas, Emma~Bou Hanna, Florian Bressand, et~al.
\newblock Mixtral of experts.
\newblock \emph{arXiv preprint arXiv:2401.04088}, 2024.

\bibitem[Jing et~al.(2021)Jing, Kong, Wang, Wang, Li, and Tan]{lts}
Ya Jing, Tao Kong, Wei Wang, Liang Wang, Lei Li, and Tieniu Tan.
\newblock Locate then segment: A strong pipeline for referring image segmentation.
\newblock In \emph{Proceedings of the IEEE/CVF Conference on Computer Vision and Pattern Recognition}, pages 9858--9867, 2021.

\bibitem[Kazemzadeh et~al.(2014)Kazemzadeh, Ordonez, Matten, and Berg]{refcoco}
Sahar Kazemzadeh, Vicente Ordonez, Mark Matten, and Tamara Berg.
\newblock Referitgame: Referring to objects in photographs of natural scenes.
\newblock In \emph{EMNLP}, 2014.

\bibitem[Kim et~al.(2022)Kim, Kim, Kwak, Lan, and Zeng]{Restr}
Namyup Kim, Dongwon Kim, Suha Kwak, Cuiling Lan, and Wenjun Zeng.
\newblock Restr: Convolution-free referring image segmentation using transformers.
\newblock In \emph{CVPR}, pages 18124--18133, 2022.

\bibitem[Kim et~al.(2024)Kim, Kang, Kim, Park, and Kwak]{risclip}
Seoyeon Kim, Minguk Kang, Dongwon Kim, Jaesik Park, and Suha Kwak.
\newblock Extending clip's image-text alignment to referring image segmentation.
\newblock In \emph{NAACL}, pages 4611--4628, 2024.

\bibitem[Lai et~al.(2023)Lai, Tian, Chen, Li, Yuan, Liu, and Jia]{lisa}
Xin Lai, Zhuotao Tian, Yukang Chen, Yanwei Li, Yuhui Yuan, Shu Liu, and Jiaya Jia.
\newblock Lisa: Reasoning segmentation via large language model.
\newblock \emph{arXiv preprint arXiv:2308.00692}, 2023.

\bibitem[Li et~al.(2022)Li, Yuan, Liang, Liu, Ji, Bai, Liu, and Bai]{can}
Bohan Li, Ye Yuan, Dingkang Liang, Xiao Liu, Zhilong Ji, Jinfeng Bai, Wenyu Liu, and Xiang Bai.
\newblock When counting meets {HMER:} counting-aware network for handwritten mathematical expression recognition.
\newblock In \emph{ECCV}, pages 197--214, 2022.

\bibitem[Lin et~al.(2014)Lin, Maire, Belongie, Hays, Perona, Ramanan, Doll{\'{a}}r, and Zitnick]{coco}
Tsung{-}Yi Lin, Michael Maire, Serge~J. Belongie, James Hays, Pietro Perona, Deva Ramanan, Piotr Doll{\'{a}}r, and C.~Lawrence Zitnick.
\newblock Microsoft {COCO:} common objects in context.
\newblock In \emph{ECCV}, pages 740--755, 2014.

\bibitem[Liu et~al.(2017)Liu, Lin, Shen, Yang, Lu, and Yuille]{liu2017}
Chenxi Liu, Zhe Lin, Xiaohui Shen, Jimei Yang, Xin Lu, and Alan~L. Yuille.
\newblock Recurrent multimodal interaction for referring image segmentation.
\newblock In \emph{ICCV}, pages 1280--1289, 2017.

\bibitem[Liu et~al.(2023{\natexlab{a}})Liu, Ding, and Jiang]{rela}
Chang Liu, Henghui Ding, and Xudong Jiang.
\newblock {GRES:} generalized referring expression segmentation.
\newblock In \emph{CVPR}, pages 23592--23601, 2023{\natexlab{a}}.

\bibitem[Liu et~al.(2023{\natexlab{b}})Liu, Ding, Zhang, and Jiang]{mmm}
Chang Liu, Henghui Ding, Yulun Zhang, and Xudong Jiang.
\newblock Multi-modal mutual attention and iterative interaction for referring image segmentation.
\newblock \emph{TPAMI}, 32:\penalty0 3054--3065, 2023{\natexlab{b}}.

\bibitem[Liu et~al.(2023{\natexlab{c}})Liu, Li, Wu, and Lee]{llava}
Haotian Liu, Chunyuan Li, Qingyang Wu, and Yong~Jae Lee.
\newblock Visual instruction tuning.
\newblock In \emph{NeurIPS}, 2023{\natexlab{c}}.

\bibitem[Liu et~al.(2023{\natexlab{d}})Liu, Ding, Cai, Zhang, Satzoda, Mahadevan, and Manmatha]{polyformer}
Jiang Liu, Hui Ding, Zhaowei Cai, Yuting Zhang, Ravi~Kumar Satzoda, Vijay Mahadevan, and R Manmatha.
\newblock Polyformer: Referring image segmentation as sequential polygon generation.
\newblock In \emph{CVPR}, pages 18653--18663, 2023{\natexlab{d}}.

\bibitem[Liu et~al.(2023{\natexlab{e}})Liu, Zhang, Qiu, Xie, Zhang, and Yao]{cris}
Sun'ao Liu, Yiheng Zhang, Zhaofan Qiu, Hongtao Xie, Yongdong Zhang, and Ting Yao.
\newblock {CARIS:} context-aware referring image segmentation.
\newblock In \emph{ACM MM}, pages 779--788, 2023{\natexlab{e}}.

\bibitem[Liu et~al.(2021)Liu, Lin, Cao, Hu, Wei, Zhang, Lin, and Guo]{swin}
Ze Liu, Yutong Lin, Yue Cao, Han Hu, Yixuan Wei, Zheng Zhang, Stephen Lin, and Baining Guo.
\newblock Swin transformer: Hierarchical vision transformer using shifted windows.
\newblock In \emph{ICCV}, pages 9992--10002, 2021.

\bibitem[Luo et~al.(2020)Luo, Zhou, Sun, Cao, Wu, Deng, and Ji]{mcn}
Gen Luo, Yiyi Zhou, Xiaoshuai Sun, Liujuan Cao, Chenglin Wu, Cheng Deng, and Rongrong Ji.
\newblock Multi-task collaborative network for joint referring expression comprehension and segmentation.
\newblock In \emph{CVPR}, 2020.

\bibitem[Mao et~al.(2016)Mao, Huang, Toshev, Camburu, Yuille, and Murphy]{refcocog}
Junhua Mao, Jonathan Huang, Alexander Toshev, Oana Camburu, Alan~L Yuille, and Kevin Murphy.
\newblock Generation and comprehension of unambiguous object descriptions.
\newblock In \emph{CVPR}, 2016.

\bibitem[Sun et~al.(2023)Sun, An, Liu, Liu, Sakaridis, Fan, and Gool]{sun2023}
Guolei Sun, Zhaochong An, Yun Liu, Ce Liu, Christos Sakaridis, Deng{-}Ping Fan, and Luc~Van Gool.
\newblock Indiscernible object counting in underwater scenes.
\newblock In \emph{CVPR}, pages 13791--13801, 2023.

\bibitem[Vaswani et~al.(2017)Vaswani, Shazeer, Parmar, Uszkoreit, Jones, Gomez, Kaiser, and Polosukhin]{transformer}
Ashish Vaswani, Noam Shazeer, Niki Parmar, Jakob Uszkoreit, Llion Jones, Aidan~N. Gomez, Lukasz Kaiser, and Illia Polosukhin.
\newblock Attention is all you need.
\newblock In \emph{NeurIPS}, pages 5998--6008, 2017.

\bibitem[Wang et~al.(2020)Wang, Wu, Zhu, Li, Zuo, and Hu]{channelattn}
Qilong Wang, Banggu Wu, Pengfei Zhu, Peihua Li, Wangmeng Zuo, and Qinghua Hu.
\newblock Eca-net: Efficient channel attention for deep convolutional neural networks.
\newblock In \emph{CVPR}, pages 11534--11542, 2020.

\bibitem[Wu et~al.(2022)Wu, Li, Li, Ding, Tong, and Tao]{rris}
Jianzong Wu, Xiangtai Li, Xia Li, Henghui Ding, Yunhai Tong, and Dacheng Tao.
\newblock Towards robust referring image segmentation.
\newblock \emph{arXiv preprint arXiv:2209.09554}, 2022.

\bibitem[Wu et~al.(2019)Wu, Kirillov, Massa, Lo, and Girshick]{detectron2}
Yuxin Wu, Alexander Kirillov, Francisco Massa, Wan-Yen Lo, and Ross Girshick.
\newblock Detectron2.
\newblock \url{https://github.com/facebookresearch/detectron2}, 2019.

\bibitem[Xia et~al.(2023)Xia, Han, Han, Pan, Song, and Huang]{gsva}
Zhuofan Xia, Dongchen Han, Yizeng Han, Xuran Pan, Shiji Song, and Gao Huang.
\newblock Gsva: Generalized segmentation via multimodal large language models.
\newblock \emph{arXiv preprint arXiv:2312.10103}, 2023.

\bibitem[Yang et~al.(2022)Yang, Wang, Tang, Chen, Zhao, and Torr]{lavt}
Zhao Yang, Jiaqi Wang, Yansong Tang, Kai Chen, Hengshuang Zhao, and Philip H.~S. Torr.
\newblock {LAVT:} language-aware vision transformer for referring image segmentation.
\newblock In \emph{CVPR}, pages 18134--18144, 2022.

\bibitem[Yu et~al.(2018)Yu, Lin, Shen, Yang, Lu, Bansal, and Berg]{mattnet}
Licheng Yu, Zhe Lin, Xiaohui Shen, Jimei Yang, Xin Lu, Mohit Bansal, and Tamara~L. Berg.
\newblock Mattnet: Modular attention network for referring expression comprehension.
\newblock In \emph{CVPR}, pages 1307--1315, 2018.

\bibitem[Zhu et~al.(2021)Zhu, Su, Lu, Li, Wang, and Dai]{deformabledetr}
Xizhou Zhu, Weijie Su, Lewei Lu, Bin Li, Xiaogang Wang, and Jifeng Dai.
\newblock Deformable {DETR:} deformable transformers for end-to-end object detection.
\newblock In \emph{ICLR}, 2021.

\end{thebibliography}
